\documentclass{article}

\PassOptionsToPackage{numbers, compress}{natbib}
\usepackage[main, final]{neurips_2026}

\usepackage[utf8]{inputenc} 
\usepackage[T1]{fontenc}    
\usepackage{hyperref}       
\usepackage{url}            
\usepackage{booktabs}       
\usepackage{amsfonts}       
\usepackage{nicefrac}       
\usepackage{microtype}      
\usepackage[table]{xcolor}  
\usepackage{tabularray}
\usepackage{multirow}
\usepackage{tabularx}  
\usepackage{makecell}
\usepackage{amsmath}
\usepackage{cleveref}
\usepackage{graphicx}
\usepackage{subcaption}
\usepackage{rotating}
\newcommand{\name}{ITO}
\usepackage{xcolor}
\definecolor{myred}{HTML}{f7b4c6}
\definecolor{myblue}{HTML}{b1cfff}
\definecolor{mygreen}{HTML}{85b87c}
\usepackage[table]{xcolor} 
\usepackage{tabularray}

\newcommand{\eg}{e.g.,\@}

\newlength\savewidth\newcommand\shline{\noalign{\global\savewidth\arrayrulewidth\global\arrayrulewidth 1pt}\hline\noalign{\global\arrayrulewidth\savewidth}}
\newcommand{\tablestyle}[2]{\setlength{\tabcolsep}{#1}\renewcommand{\arraystretch}{#2}\centering\footnotesize}

\title{ITO: Multi-View Alignment and Training-Time Fusion for Image-Text Pretraining}

\author{%
  Hanpeng Liu$^{1,2}$\thanks{Work done as a research intern at Li Auto Inc.} \quad
  Yaqian Li$^{2}$\thanks{Corresponding authors.} \quad
  Zidan Wang$^{1}$ \quad
  Shuoxi Zhang$^{3}$ \quad
  Zonglin Zhao$^{1}$ \\
  \bfseries
  Zihao Bo$^{2}$ \quad
  Rinyoichi Takezoe$^{2}$ \quad
  Kaiwen Long$^{2}$ \quad
  Kun He$^{1}$\footnotemark[2] \\[4pt]
  \normalfont
  $^{1}$School of Computer Science and Technology, \\
  Huazhong University of Science and Technology, Wuhan, China \\
  $^{2}$Li Auto Inc. \\
  $^{3}$Institute of AI for Industries, Chinese Academy of Sciences \\
  \texttt{liyaqian@lixiang.com}, \texttt{brooklet60@hust.edu.cn}
}

\begin{document}

\maketitle

\begin{abstract}
Image--text contrastive pretraining has become a dominant paradigm for visual representation learning, yet existing methods often yield representations that remain partially organized by modality rather than by semantics. We propose \textbf{ITO}, a framework addressing this limitation through two complementary mechanisms with distinct roles. \emph{Multimodal multiple alignment} enriches supervision by constructing diverse cross-modal correspondences from multi-view image augmentations, providing the primary source of discriminative gain. A lightweight \emph{training-time multimodal fusion} module then acts as a training-time regularizer, encouraging the encoders to produce features that are compatible under fusion. Crucially, the fusion module is discarded at inference, preserving the efficiency of standard dual-encoder architectures while incurring training-time overhead only. Extensive experiments across pretraining scales from millions to billions of image--text pairs show that ITO outperforms strong contrastive baselines at moderate scale and yields consistent gains over CLIP under identical data, backbone, and compute at the 100M--1B scale, on classification, retrieval, and multimodal benchmarks. Controlled comparisons show that these gains are not explained by additional training compute or image augmentation alone. Our analysis further reveals that the contribution of fusion grows with data scale and that it stabilizes optimization, mitigating the late-stage overfitting commonly observed in aggressive contrastive learning. Code is available at \url{https://github.com/showstarpro/ITO}.
\end{abstract}
\section{Introduction}

Recent foundation models such as CLIP~\cite{clip,openclip} have fundamentally reshaped visual representation learning through large-scale image--text contrastive pretraining, demonstrating strong transferability across zero-shot classification, retrieval, and as visual backbones for multimodal large language models~\cite{llava15}. A growing body of follow-up work~\cite{SLIP,alignclip,xiao2024flair} has further strengthened alignment quality and scaling behavior through improved data curation, alternative objectives, and architectural refinements, establishing image--text contrastive learning as a dominant paradigm for large-scale visual pretraining.

Despite this success, we identify a fundamental limitation that has received insufficient attention: \emph{strong pairwise alignment does not imply that the representation space is organized by semantics rather than by modality.} Contrastive objectives optimize instance-level matching between paired images and texts, but place no explicit constraint on the global geometry of the resulting embedding space. As a consequence, image and text embeddings can remain organized by modality even when retrieval and classification metrics are strong---a tendency we term \emph{modality-induced separation}: cross-modal semantic neighbors (an image and a text of the same concept) are not reliably closer than cross-modal non-neighbors. This notion is related to but distinct from the \emph{modality gap}~\cite{liang2022mind}, a mean offset between the two modality clusters that can persist even when semantics are well organized, and from embedding anisotropy, which is a unimodal property of how features spread. We emphasize that reducing modality-induced separation does not mean collapsing the two modalities into a single geometric cluster; rather, it means organizing the shared embedding space so that semantically similar samples are close regardless of modality, while unrelated samples remain well separated.

\begin{figure}[t!]
\centering
\includegraphics[width=0.85\linewidth]{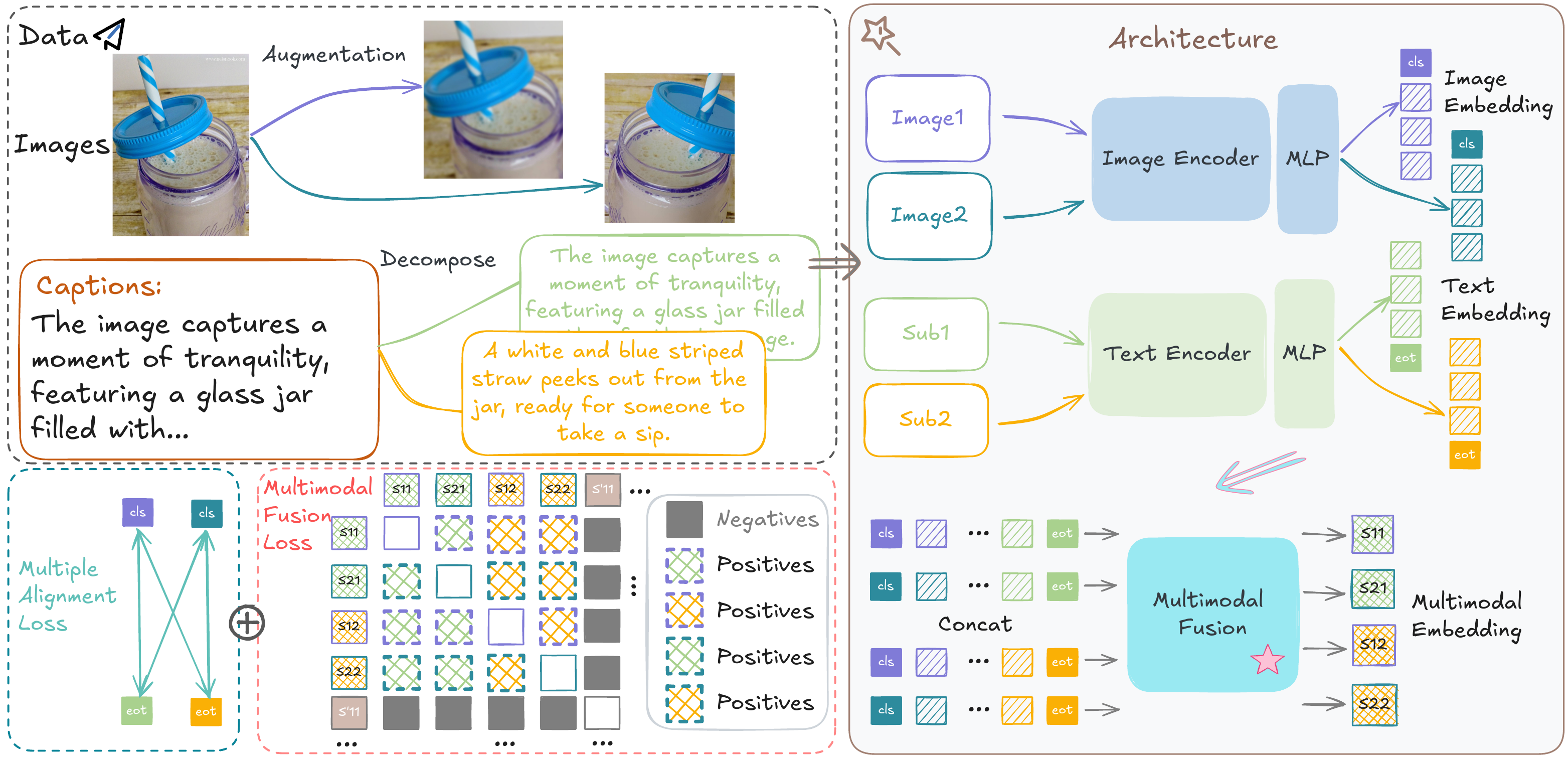}
\caption{
\textbf{Overview of the proposed \name\ training framework.}
Starting from standard image–text contrastive pretraining, \name\ restructures supervision through multimodal multiple alignment and introduces a lightweight multimodal fusion module during training. Multiple augmented image–text pairs derived from the same sample are used to enrich instance-level alignment, while training-time fusion encourages encoders to produce features that are compatible under fusion, reducing modality-induced separation. Importantly, the fusion module is used only during training and is discarded at inference time, allowing \name\ to retain a standard dual-encoder architecture for efficient deployment.}
\label{fig:motivation}
\vspace{-2.5em}
\end{figure}

This distinction---between \emph{aligning} paired instances and \emph{organizing} the representation space by semantics---is the central concern of this work. Prior work on cross-modal fusion~\cite{FIBER,alignclip,xiao2024flair} has explored richer multimodal interaction, but as part of the inference architecture, conflating the goal of improving representation structure with the cost of modifying the inference pipeline. A cleaner approach uses cross-modal interaction purely as a \emph{training signal} to reshape the dual encoders, and then discards it at inference; to our knowledge, this has not been systematically explored.

This motivates \textbf{ITO}, an image--text contrastive pretraining framework built around a unified principle: \emph{instance-level alignment and global representation organization require distinct, dedicated mechanisms.} ITO addresses each through a dedicated component. First, \emph{Multimodal multiple alignment} restructures the cross-modal supervision signal itself: rather than pairing each image with a single caption, it constructs many-to-one (or many-to-many) correspondences from multiple augmented image views against the same caption within each batch. This densified supervision is the primary engine of discriminative gain. Second, a lightweight \emph{training-time multimodal fusion} module acts as a training-time regularizer on the encoder representations: by applying a contrastive objective over fused multimodal tokens, it propagates gradients back to the dual encoders, which must then produce image and text tokens that remain discriminative and mutually consistent after cross-modal attention. Crucially, the fusion module is discarded entirely at inference, so ITO retains the full efficiency and scalability of a standard dual-encoder architecture. This makes ITO a drop-in replacement in settings where inference cost is the binding constraint, such as large-scale image--text retrieval and embedding serving, where cross-modal interaction at query time is prohibitive but extra one-off training compute is acceptable. The overall framework is illustrated in~\Cref{fig:motivation}.

The distinguishing features of ITO relative to prior art are two-fold. First, whereas SLIP~\cite{SLIP} introduces intra-modal visual augmentation as an auxiliary objective and LaCLIP~\cite{LaCLIP}/VeCLIP~\cite{VeCLIP} expand textual views, all while preserving the one-to-one cross-modal pairing structure, ITO restructures the cross-modal supervision itself through many-to-one correspondences---a qualitatively different form of enrichment. Second, whereas inference-time fusion architectures such as ALBEF~\cite{DBLP:conf/nips/LiSGJXH21}, FLAIR~\cite{xiao2024flair}, and FIBER~\cite{FIBER} couple cross-modal interaction with deployment, and knowledge distillation~\cite{hinton2015distilling} requires a separately trained teacher, ITO uses fusion exclusively as a training-time regularizer with no inference-time role and no teacher, separating the goal of representation shaping from the constraint of inference efficiency.

Our experiments reveal a clear division of labor. Multi-view alignment contributes the dominant share of discriminative gain across all data scales, and this gain is not explained by extra compute or augmentation: CLIP trained with twice the compute recovers less than one point of the gap, and ITO exceeds SLIP, which uses multiple image views under one-to-one supervision (\Cref{sec:compute}). Training-time fusion plays a complementary role whose direct accuracy contribution is modest; we therefore ground it in optimization evidence rather than in a geometric claim: its gain grows with data scale, it suppresses the late-stage overfitting that alignment-only training still exhibits, and its effect is non-monotone in the fusion weight, as expected of a regularizer. Extensive evaluations across zero-shot classification~\cite{objectnet,cimpoi14describing,veeling2018rotation,inske}, image--text retrieval, and multimodal benchmarks confirm these gains. They come at a non-trivial training-time cost (\Cref{tab:cost}), but with no inference-time overhead.

\section{Related Work}
\label{relate}

\textbf{Image--Text Contrastive Learning.} CLIP~\cite{clip} established image--text contrastive learning as a dominant paradigm for visual representation learning, with follow-up work improving data curation, training objectives, and scaling behavior~\cite{siglip,metaclip,cyclip,superclip}. The visual encoders learned this way transfer effectively to zero-shot classification, retrieval, and as backbones for multimodal large language models~\cite{DBLP:conf/cvpr/GoyalKSBP17}. Despite this success, most approaches focus on instance-level pairwise alignment and do not explicitly constrain how representations are globally organized in the shared embedding space, motivating recent efforts to examine deeper structural properties of learned representations.

\textbf{Strengthening Alignment via View Augmentation.} A line of work enriches contrastive supervision by augmenting individual modality views. SLIP~\cite{SLIP} adds intra-modal visual augmentations as an auxiliary SimCLR-style objective, but the cross-modal contrastive signal still pairs each image with one caption. LaCLIP~\cite{LaCLIP}, VeCLIP~\cite{VeCLIP}, and DreamLIP~\cite{dreamlip} expand textual views using large language or vision--language models, again preserving instance-level one-to-one cross-modal supervision. CyCLIP~\cite{cyclip} adds geometric consistency constraints on top of CLIP's pairing without altering the supervision structure. Multi-crop and multi-positive objectives~\cite{simclr,mocov3} draw their positives from within a modality (crop $\leftrightarrow$ crop of the same image) or from existing pairs, and densify single-modality structure. In contrast, our \emph{multi-view alignment} restructures the \emph{cross-modal} supervision itself: multiple augmented image views are paired against the same caption inside the image--text InfoNCE to form many-to-one (or, optionally, many-to-many) cross-modal correspondences within each batch. This is a change to the supervision structure rather than to the augmentation set, and we show in~\Cref{sec:compute} that its benefit is not reproduced by adding image views under one-to-one supervision.

\textbf{Cross-Modal Fusion and Auxiliary Training Signals.} Another line of work introduces cross-modal interaction during pretraining. ALBEF~\cite{DBLP:conf/nips/LiSGJXH21}, FIBER~\cite{FIBER}, and related architectures incorporate cross-attention or joint encoding modules that remain active at inference time. AlignCLIP~\cite{alignclip} reduces the modality gap~\cite{liang2022mind} by sharing parameters across encoders, while FLAIR~\cite{xiao2024flair} uses text-conditioned pooling that modifies the inference-time encoder behavior. In all of these, fusion is tightly coupled with the inference architecture or with task-specific designs. Knowledge distillation~\cite{hinton2015distilling} also shapes a deployed model with an auxiliary network, but requires a separately trained teacher whose outputs the student imitates. ITO instead uses a fusion module \emph{exclusively as a training signal}, with neither an inference-time role nor a teacher: gradients from a multimodal contrastive objective on fused tokens propagate back to reshape the dual encoders, after which the fusion module is discarded. To our knowledge, this combination---cross-modal multi-view supervision with a train-only fusion regularizer---has not been explored in prior image--text contrastive learning, and it allows ITO to retain a standard CLIP-style dual-encoder architecture at deployment.

\section{Method}
\subsection{Preliminaries: CLIP-style Contrastive Pretraining}
We build upon the standard image–text contrastive learning framework introduced by CLIP. Given a batch of 
$B$ image–text pairs $\{(I^{n}, T^{n})\}_{n=1}^{B}$
, an image encoder $\textit{E}_I$ and a text encoder $\textit{E}_T$ are used to extract visual and textual representations, which are then projected into a shared embedding space via projection heads $P_I$ and $P_T$:
\begin{equation}
    Y^n = P_I(E_I(I^n)), \ \ \  \ \ \  Z^n= P_T(E_T(T^n)).
\end{equation}
Image–text alignment is achieved through a symmetric InfoNCE objective. Specifically, the image-to-text and text-to-image losses are defined as
\begin{subequations}
\label{clip_losses}
\begin{align}
\mathcal{L}_{I \to T}
&= - \sum_{n=1}^{B} 
\log \frac{
\exp\big( \langle Y^n , Z^n \rangle / \tau \big)
}{
\sum_{m=1}^{B} 
\exp\big( \langle Y^n , Z^m \rangle / \tau \big)
}, \label{clip_image_text} \\
\mathcal{L}_{T \to I}
&= - \sum_{n=1}^{B} 
\log \frac{
\exp\big( \langle Z^n , Y^n \rangle / \tau \big)
}{
\sum_{m=1}^{B} 
\exp\big( \langle Z^n , Y^m \rangle / \tau \big)
}. \label{clip_text_image}
\end{align}
\end{subequations},
where $\left\langle \cdot \right\rangle$ denotes cosine similarity and $\tau$ is a learnable temperature. The overall CLIP loss is given by
\begin{equation}
    \mathcal{L}_{CLIP} = \frac{1}{2} (\mathcal{L}_{I \to T} +\mathcal{L}_{T \to I}).
\end{equation}
This dual-encoder formulation enables efficient and scalable inference, which we preserve throughout this work.

\subsection{Multimodal Multiple Alignment}
To enrich contrastive supervision beyond one-to-one image–text pairing, we introduce Multimodal multiple alignment, which restructures the supervision signal within a training batch. Instead of treating each image–text pair as a single positive instance, this strategy exposes the model to multiple image–text correspondences derived from the same underlying sample, enabling more flexible instance-level alignment.

Concretely, for each original image–text pair $(I_i^n, T_j^n)$, multiple image–text combinations are constructed through standard image and text perturbations, resulting in a set of augmented pairs $\{(I_i^n, T_j^n)\}$. By default, we use two image views ($i \in \{1,2\}$) and a single text view ($j = 1$)---the configuration we recommend for all large-scale experiments. A variant with two text views ($j \in \{1,2\}$), denoted \textit{ITO\_sub2}, is helpful only at moderate data scales. We present the formulation under \textit{ITO\_sub2}; the default is the special case $j=1$.

For each augmented image–text pair $(I^n_i, T^n_j)$, we compute a bidirectional contrastive loss following the CLIP formulation, including image-to-text and text-to-image directions. Importantly, we retain the standard batch-wise negative sampling strategy: the corresponding augmented pair is treated as the positive sample, while all other samples within the batch serve as negatives. This process is repeated for all valid combinations of image and text views, and the final alignment loss is computed as the average over these losses. Formally, the multiple alignment loss is defined as
\begin{equation}
    \mathcal{L}_{\mathrm{Align}} = \frac{1}{4}\sum_{i=1}^{2}\sum_{j=1}^{2} \frac{1}{2}\left[ \mathcal{L}_{I_i \to T_j} + \mathcal{L}_{T_j \to I_i} \right],
\end{equation}
where $\mathcal{L}_{I_i \to T_j}$ and $\mathcal{L}_{T_j \to I_i}$ denote the image-to-text and text-to-image InfoNCE losses, respectively.

By increasing the diversity and number of positive image–text pairings within each batch, Multimodal multiple alignment enriches instance-level supervision and improves alignment robustness without introducing additional inference-time cost. Note that the multiplicity enters the \emph{cross-modal} objective: each image view is a positive for the shared caption, rather than for another crop of the same image as in multi-crop self-supervised learning. However, this strategy alone still scores each image and text independently, and never requires their features to remain consistent under joint processing, which motivates the training-time fusion objective below.

\subsection{Training-Time Multimodal Fusion}
\label{sec:fusion}
While multiple alignment strengthens correspondence between individual image–text pairs, it scores the two modalities only through a dot product of pooled embeddings. To add a complementary training signal, we introduce a lightweight training-time multimodal fusion module that acts as a regularizer on the encoders' representations.

Given an augmented image–text pair $(I^n_i, T^n_j)$, we obtain their corresponding visual tokens $Y^n_i$ and textual tokens $Z^n_j$. These tokens are concatenated to form a joint multimodal sequence:
\begin{equation}
H^n_{i,j} = \mathrm{Concat}(Y^n_i, Z^n_j),
\end{equation}
where $\mathrm{Concat}(\cdot,\cdot)$ denotes token-wise concatenation. The joint sequence is then processed by a lightweight fusion module $F_M$, implemented as a two-layer Transformer with bidirectional self-attention over the joint sequence, producing fused multimodal tokens:
\begin{equation}
    S^n_{i,j} = F_M(H^n_{i,j}).
\end{equation}
We use the output at the first position of the joint sequence, i.e., the image \texttt{[CLS]} token, as the fused multimodal representation. Since the vision tokens come first and have a fixed length, this is a fixed read-out position that, under bidirectional self-attention, attends over the entire joint sequence, and it reuses the aggregation role \texttt{[CLS]} already plays in the ViT image encoder. We compare it with alternative read-outs in Appendix~\ref{app:readout}.

During training, the fusion objective encourages consistency among multimodal representations derived from the same underlying image–text pair, while pushing apart representations from different pairs. Specifically, fused representations obtained from different augmentations of the same image–text pair are treated as positive samples, while fused representations from other samples within the batch serve as negatives. We adopt a contrastive formulation with multiple positives to account for this structure. The loss for a fused representation is defined as:
\begin{equation}
\scalebox{0.95}{$
\mathcal{L}_{S^n_{1,1}  } = - \log{\frac{\exp[\left \langle S^n_{1,1} \cdot S^n_{2,1} \right \rangle / \tau ] + \sum_{i=1}^{2}\exp[\left \langle S^n_{1,1} \cdot S^n_{i,2} \right \rangle / \tau ]  }{\sum^B_{m=1} \sum^2_{i=1} \sum^2_{k=1} \exp[\left \langle S^n_{1,1} \cdot S^m_{i,k} \right \rangle / \tau ]}},
$}
\end{equation}
where the trivial self-pair $(m=n,i=1,k=1)$ is excluded. The overall fusion loss is:
\begin{equation}
    \mathcal{L}_{\mathrm{Fusion}} = \frac{1}{4B} \sum_{n=1}^{B} \sum_{i=1}^{2} \sum_{j=1}^{2} \mathcal{L}_{S^n_{i,j}}.
\end{equation}

\textbf{Gradient path.} The fusion module receives the image and text tokens without stop-gradient, and the fused \texttt{[CLS]} representation is a self-attention function of both. The fusion loss therefore backpropagates through $F_M$ into both encoders. Because $F_M$ has no private input path, it cannot reduce $\mathcal{L}_{\mathrm{Fusion}}$ on its own: the encoders must produce image and text tokens that remain discriminative and mutually consistent after cross-modal self-attention. In this way $\mathcal{L}_{\mathrm{Fusion}}$ acts as a training-time regularizer: it encourages features that are not only linearly separable (as in vanilla contrastive learning) but also compatible under fusion, while preserving the dual-encoder architecture. We characterize its effect through optimization behavior rather than through a claim about global geometry (\Cref{align_integration}).


\subsection{Overall Objective and Inference}

The final training objective combines the multiple alignment loss and the multimodal fusion loss:
\begin{equation}
\mathcal{L} = \mathcal{L}_{\mathrm{Align}} + \lambda \mathcal{L}_{\mathrm{Fusion}}.
\end{equation}
Here, $\lambda$ balances the trade-off between discriminative intensity (from alignment) and fusion regularization.
At inference time, ITO reduces to a standard dual-encoder model identical to CLIP, with no fusion modules and no additional inference-time overhead, enabling direct replacement of existing image--text contrastive encoders. The improved representation quality, however, comes at a non-trivial training-time cost (approximately $2.2\times$ training time and $2.1$--$2.3\times$ peak memory compared to OpenCLIP), reported in~\Cref{tab:cost}.
\section{Experiments}
\subsection{Implementation Details}
\label{settings}
\textbf{Datasets.} We evaluate our method across image–text datasets spanning from millions to billions of samples in order to assess effectiveness, robustness, and scalability.
Specifically, we conduct pretraining on CC3M~\cite{cc3m}, CC12M~\cite{cc12m}, and YFCC15M~\cite{orralbaE11}, which are widely used benchmarks for image--text contrastive pretraining.
To study large-scale behavior, we further perform experiments on Laion100M (a 100M subset of Laion400M~\cite{laion400m}) and the billion-scale DataComp-1B~\cite{datacomp-1b} dataset.
Due to computational constraints, large-scale baselines are limited: on Laion100M we compare against CLIP and SLIP, and on DataComp-1B only against CLIP. Billion-scale reruns of SLIP, SigLIP, and FLAIR are beyond our budget, and public checkpoints trained on a matched corpus are not available. At these scales, CLIP therefore serves as a controlled reference that holds data, backbone, and schedule fixed.
Detailed dataset statistics are provided in appendix~\Cref{datasets}.

\textbf{Data Augmentation.}
We adopt standard image augmentations following prior work (\eg MoCo v3~\cite{mocov3} and SLIP~\cite{SLIP}) to generate two image views per sample; the exact augmentation policy is given in Appendix~\ref{app:aug}. For text, we consider two settings: a default configuration using a single caption, and an optional variant that samples two sub-descriptions from the original text. The latter is denoted as \name\_sub2. Sub-description sampling (\name\_sub2) is applied only to CC3M, CC12M, YFCC15M, and DataComp-1B experiments; sub-descriptions are generated following FLAIR~\cite{xiao2024flair} (Appendix~\ref{app:sub2}). These augmentations are used solely to construct multiple views for contrastive supervision.

\textbf{Pretraining Settings.}
All models are implemented based on OpenCLIP~\cite{openclip}.
Unless otherwise specified, we use ViT-B/16 as the vision encoder and the standard CLIP text encoder.
Images are resized to $224{\times}224$, and text sequences are tokenized to a maximum length of 77 tokens.
Models are optimized using AdamW with a cosine learning rate schedule and linear warmup.
We train models for 30 epochs on CC3M, CC12M, YFCC15M and Laion100M.
For DataComp-1B, models are trained for one epoch due to computational cost; an additional 10-epoch ViT-B/16 model is trained to study data scaling effects. 
Batch size and learning rate are adjusted according to the dataset scale.
The fusion loss weight $\lambda$ is set to 2 by default. All other hyperparameters follow OpenCLIP defaults.
All training is conducted on NVIDIA A800 GPUs. To ensure fair comparison, we reproduce these methods using the versions of the datasets we have downloaded. 

\textbf{Evaluation Protocol.}
We evaluate pretrained models using only the image and text encoders, without any fusion modules, to assess representation quality under a standard dual-encoder setting.
Four categories of downstream tasks are considered.

\textit{Zero-shot Image Classification.}
We follow the evaluation protocol of EVA-CLIP~\cite{DBLP:journals/corr/abs-2303-15389} and report top-1 accuracy on 26 datasets covering generic, fine-grained, and domain-specific classification tasks.


\textit{Linear Image Classification \& Retrieval.} We perform linear probing on frozen visual features and bidirectional zero-shot retrieval on COCO~\cite{DBLP:conf/eccv/LinMBHPRDZ14}, Flickr30K~\cite{DBLP:conf/iccv/PlummerWCCHL15}, and DOCCI~\cite{DBLP:conf/eccv/OnoeRBBCGKPPTWB24} (Recall@1/5/10).

\textit{Vision-Language Understanding.}
To assess transferability to multimodal systems, we follow the official LLaVA-1.5~\cite{llava15} evaluation protocol on 13 benchmarks, including VQAv2~\cite{vqav2}, MM-Vet~\cite{DBLP:conf/icml/YuYLWL0WW24}, POPE~\cite{DBLP:conf/emnlp/LiDZWZW23}, and MMStar~\cite{DBLP:conf/nips/ChenLDZZCDWQLZ24}.
These experiments evaluate the pretrained visual encoders as backbones rather than introducing new VLM architectures.

All baselines are reproduced and evaluated under the same pipeline for fair comparison and reproducibility.

\begin{table}[t]
\centering
\caption{Average top-1 zero-shot classification accuracy across 26 public benchmarks following the EVA-CLIP~\cite{DBLP:journals/corr/abs-2303-15389}  protocol. All models use ViT-B/16 and are trained for 30 epochs unless otherwise noted. The upper block compares ITO against contrastive baselines under the same backbone and training schedule; the lower block reports scaling behavior with longer training (10 epochs) and a larger backbone (ViT-L/16) on DataComp-1B. Due to computational constraints, SigLIP and FLAIR are evaluated only on smaller datasets and SLIP up to Laion100M; CLIP is the only baseline run consistently across all pretraining scales, and at large scale the comparison is against CLIP under identical data, backbone, and compute. Per-benchmark results are provided in Appendix~\Cref{tab::datacomp_b10}. Best results are in \textbf{bold}.}
\label{tab::classification}
\setlength{\tabcolsep}{6pt}
\renewcommand{\arraystretch}{1.05}
\small
\begin{tabular}{lcccccc}
\toprule
Pretraining Data & CLIP & SLIP & SigLIP & FLAIR & ITO & ITO\_sub2 \\
\midrule
CC3M        & 19.3 & 21.4 & 19.3 & 23.2 & \textbf{24.9} & \textbf{24.9} \\
CC12M       & 34.1 & 36.0 & 35.0 & 35.5 & 38.4 & \textbf{40.4} \\
YFCC15M     & 30.5 & --   & --   & --   & 34.5 & \textbf{35.4} \\
Laion100M   & 53.5 & 51.3 & --   & --   & \textbf{56.1} & --   \\
DataComp-1B & 54.2 & --   & --   & --   & 56.6 & \textbf{57.0} \\
\midrule
\multicolumn{7}{l}{\textit{Scaling: longer training and larger backbones on DataComp-1B}} \\
\midrule
\quad ViT-B/16, 10 epochs & 63.7 & -- & -- & -- & \textbf{64.9} & 63.2 \\
\quad ViT-L/16, 1 epoch   & 60.7 & -- & -- & -- & \textbf{63.0} & 61.5 \\
\bottomrule
\end{tabular}
\vspace{-2em}
\end{table}

\subsection{Zero-shot Image Classification}
\label{sec:zs}

We evaluate zero-shot image classification on 26 public benchmarks following the EVA-CLIP~\cite{DBLP:journals/corr/abs-2303-15389}. Average top-1 accuracy is reported in~\Cref{tab::classification}; full per-benchmark results are deferred to Appendix~\Cref{tab::datacomp_b10}.

At moderate scale (CC3M, CC12M, YFCC15M), ITO outperforms CLIP and prior contrastive baselines including SLIP~\cite{SLIP}, SigLIP~\cite{siglip}, and FLAIR~\cite{xiao2024flair}. At the 100M and 1B scales, where CLIP is the only baseline run at every scale (SLIP is additionally available on Laion100M), ITO yields consistent gains over CLIP under identical data, backbone, and compute. The benefits of multi-view alignment and training-time fusion are thus not confined to a single data regime, although we do not claim superiority over stronger baselines at billion scale.

We further analyze the effect of textual diversity through sub-description sampling. \name\_sub2 is not part of the default ITO, and the headline results do not depend on it. Its behavior reflects the data regime. On CC3M, CC12M, and YFCC15M, where native captions are short and sparse, sub-descriptions inject textual diversity the captions lack, and \name\_sub2 matches or exceeds the default ITO. On DataComp-1B, where native captions are already lexically rich, sub-descriptions become largely redundant paraphrases and act as noise: \name\_sub2 falls slightly below both ITO and CLIP at 10 epochs (63.2 vs.\ 63.7 for CLIP). We therefore adopt the default ITO configuration for all large-scale experiments. The scaling block of ~\Cref{tab::classification} further shows that longer training (10 epochs) and a larger backbone (ViT-L/16) both continue to improve ITO, with the gap to CLIP preserved or widened.

\subsection{Linear Image Classification}
\label{sec:linear}

\begin{table}[t]
\centering
\caption{Linear probing top-1 accuracy of ITO and CLIP across pretraining scales. We report ImageNet-1K (IN-1k) and the average over five datasets (IN-1k, Flowers-102, Stanford Cars, Pets, Food-101). Per-dataset results, additional baselines, and ITO\_sub2 variants are provided in Appendix~\Cref{linear_full}. Best results in each block are in \textbf{bold}.}
\label{tab::linear-recap}
\setlength{\tabcolsep}{5pt}
\renewcommand{\arraystretch}{1.1}
\small
\begin{tabular}{lcccccccc}
\toprule
& \multicolumn{2}{c}{YFCC15M} & \multicolumn{2}{c}{Laion100M} & \multicolumn{2}{c}{DataComp-1B (1ep)} & \multicolumn{2}{c}{DataComp-1B (10ep)} \\
\cmidrule(lr){2-3} \cmidrule(lr){4-5} \cmidrule(lr){6-7} \cmidrule(lr){8-9}
Method & IN-1k & Avg & IN-1k & Avg & IN-1k & Avg & IN-1k & Avg \\
\midrule
CLIP & 51.82 & 55.88 & 67.27 & 81.86 & 67.39 & 82.14 & 74.99 & 87.62 \\
ITO  & \textbf{60.52} & \textbf{63.76} & \textbf{68.70} & \textbf{83.56} & \textbf{69.87} & \textbf{82.53} & \textbf{76.20} & \textbf{89.18} \\
\bottomrule
\end{tabular}
\vspace{-2em}
\end{table}

We further evaluate the quality of learned visual representations via linear probing. Following standard practice, we train a linear classifier on top of frozen visual features for 30 epochs using AdamW with a cosine learning-rate schedule. Top-1 accuracy on ImageNet-1K and the average over five classification datasets are reported in~\Cref{tab::linear-recap} for medium- to billion-scale pretraining; results for CC3M and CC12M are visualized in Appendix~\Cref{fig:linear_cc3m} and ~\Cref{fig:linear_cc12m}, with full per-dataset numbers and additional baselines deferred to the appendix.

Across all pretraining scales, ITO consistently outperforms CLIP under linear evaluation, indicating improved linear separability of the learned visual representations. The gap is largest at medium scale (e.g., +7.9 average points on YFCC15M) and remains consistent at billion scale (+1.6 points on DataComp-1B with 10-epoch training). Importantly, the linear-probing advantage holds even though the fusion module is discarded at inference, showing that ITO training changes the visual encoder itself rather than relying on an inference-time helper. On DataComp-1B, longer training raises ITO from 82.53 (1 epoch) to 89.18 (10 epochs) while preserving the gain over CLIP; full results are in the appendix.

\subsection{Zero-shot Image-Text Retrieval}
\label{sec:retrieval}

\begin{table}[t]
\centering
\caption{Zero-shot image-text retrieval (Recall@1) on MSCOCO and Flickr30k. CC12M and DataComp-1B (ViT-B/16, 1 epoch) represent medium- and large-scale pretraining respectively. Comparisons against additional baselines (SLIP, SigLIP, FLAIR), all five pretraining scales, and full Recall@5/10 metrics are provided in Appendix~\Cref{tab::retrieval-full}. Best results are in \textbf{bold}.}
\label{tab::retrieval}
\setlength{\tabcolsep}{4pt}
\renewcommand{\arraystretch}{1.1}
\small
\begin{tabular}{lcccccccc}
\toprule
& \multicolumn{4}{c}{CC12M} & \multicolumn{4}{c}{DataComp-1B (1 epoch)} \\
\cmidrule(lr){2-5} \cmidrule(lr){6-9}
& \multicolumn{2}{c}{MSCOCO} & \multicolumn{2}{c}{Flickr30k} & \multicolumn{2}{c}{MSCOCO} & \multicolumn{2}{c}{Flickr30k} \\
\cmidrule(lr){2-3} \cmidrule(lr){4-5} \cmidrule(lr){6-7} \cmidrule(lr){8-9}
Method & I$\to$T & T$\to$I & I$\to$T & T$\to$I & I$\to$T & T$\to$I & I$\to$T & T$\to$I \\
\midrule
CLIP     & 34.17 & 23.04 & 62.23 & 45.74 & 47.08 & 29.51 & 72.68 & 54.73 \\
ITO      & 42.30 & 29.62 & \textbf{72.49} & 54.50 & 49.26 & 31.30 & \textbf{75.94} & 56.79 \\
ITO\_sub2 & \textbf{43.94} & \textbf{30.51} & 72.29 & \textbf{56.65} & \textbf{49.50} & \textbf{31.89} & 73.37 & \textbf{56.79} \\
\bottomrule
\end{tabular}
\vspace{-2em}
\end{table} 
We evaluate zero-shot image-text retrieval to assess cross-modal alignment quality on MSCOCO~\cite{DBLP:conf/eccv/LinMBHPRDZ14} and Flickr30k~\cite{DBLP:conf/iccv/PlummerWCCHL15}. Bidirectional Recall@1 is reported in ~\Cref{tab::retrieval}; results across all five pretraining scales, full Recall@5/10, and fine-grained retrieval on DOCCI~\cite{DBLP:conf/eccv/OnoeRBBCGKPPTWB24} are provided in Appendix Tables~\ref{tab::retrieval-full} and~\Cref{tab::fine-grained}.

Two findings are consistent across all retrieval settings. First, ITO outperforms CLIP, SigLIP, and FLAIR on both retrieval directions and both datasets, indicating stronger cross-modal alignment under the standard dual-encoder protocol. Second, the role of textual diversity differs sharply across data scales: at medium scale (CC12M), sub-description sampling (ITO\_sub2) yields additional gains of 1.6--2.2 R@1 points, whereas at billion scale this advantage attenuates as the abundant native variation in web captions already saturates the textual supervision signal. Since retrieval depends directly on cross-modal similarity, these results indicate that the benefits of ITO extend beyond same-modality discriminability (Section~\ref{sec:linear}) to cross-modal alignment.

\subsection{Transfer to MLLM Benchmarks}
\begin{table}[t]
\centering
\caption{The performance of \name~on a broad range of multimodal tasks. The best results are \textbf{bold}.}
\label{llava}
\resizebox{0.9\linewidth}{!}{
\begin{tblr}{
  cell{1}{15} = {b},
  cell{1}{3} = {b},
  cell{1}{4} = {b},
  cell{1}{5} = {b},
  cell{1}{6} = {b},
  cell{1}{7} = {b},
  cell{1}{8} = {b},
  cell{1}{9} = {b},
  cell{1}{10} = {b},
  cell{1}{11} = {b},
  cell{1}{12} = {b},
  cell{1}{13} = {b},
  cell{1}{14} = {b},
  cell{2}{1} = {r=2}{},
  cell{4}{1} = {r=2}{},
  hline{1, 6}={-}{1pt},
  hline{2,4}={-}{},
}
Dataset & Method   & \begin{sideways}VQAv2\end{sideways} & \begin{sideways}GQA\end{sideways} & \begin{sideways}T-VQA\end{sideways} & \begin{sideways}SciQA-I\end{sideways} & \begin{sideways}VisWiz\end{sideways} & \begin{sideways}MMB-en\end{sideways} & \begin{sideways}MMB-cn\end{sideways} & \begin{sideways}MMVet\end{sideways} & \begin{sideways}POPE-r\end{sideways} & \begin{sideways}POPE-p\end{sideways} & \begin{sideways}POPE-a\end{sideways} & \begin{sideways}MMMU\end{sideways} & \begin{sideways}MMStar\end{sideways} \\
100M&CLIP & 66.53                               & 56.36                             & 47.55                                 & 65.44                                   & 36.80                                 & 53.69                                & 46.13                                & 18.30                                & 84.09                                & 82.90                                 & 76.40                                 & 33.60                               & 28.87                                \\
&\textbf{ITO}     & \bf68.47                               & \bf57.23                             & \bf48.71                                 & \bf66.29                                   & \bf44.25                                & \bf54.12                                & \bf46.65                                & \bf19.80                                & \bf84.74                                & \bf83.63                                & \bf77.30                                 & \bf33.90                               &\bf29.13    \\
1B & CLIP & 70.42 & 57.93 & 50.24 & 65.00 & \bf45.28 & 48.45 & 55.67 & 18.40 & 83.63 & 82.71 & 78.63 & \bf34.00 &  29.33 \\
& \textbf{ITO} & \bf73.19 & \bf59.99 & \bf50.89 & \bf66.24 & 42.92 & \bf50.52 & \bf58.85 & \bf21.90 & \bf85.46 & \bf84.23 & \bf80.46 & 33.90 & \bf31.13 \\   

\end{tblr}
}
\vspace{-1em}
\end{table}

Many multimodal large language models rely on a pretrained vision encoder as a fixed perceptual backbone.
To evaluate the transferability of learned visual representations, we follow the LLaVA-1.5~\cite{llava15} protocol and integrate Vicuna-7B~\cite{vicuna2023} with vision encoders pretrained using CLIP and \name, all based on ViT-B/16.
Unless otherwise specified, the vision encoder is frozen during multimodal fine-tuning, isolating the effect of visual pretraining.

We evaluate models pretrained on Laion100M and on the 10-epoch version of DataComp-1B across 13  Multimodal Large Language Model benchmarks, including VQAv2~\cite{vqav2}, GQA~\cite{DBLP:conf/emnlp/AinslieLJZLS23}, T-VQA~\cite{DBLP:conf/cvpr/SinghNSJCBPR19}, SciQA~\cite{DBLP:conf/nips/LuMX0CZTCK22}, VizWiz~\cite{DBLP:conf/cvpr/Gurari0SGLGLB18}, MMBench~\cite{DBLP:conf/eccv/LiuDZLZZYWHLCL24}, MMVet~\cite{DBLP:conf/icml/YuYLWL0WW24}, POPE~\cite{DBLP:conf/emnlp/LiDZWZW23}, MMMU~\cite{DBLP:conf/cvpr/YueNZ0LZSJRSWYY24}, and MMStar~\cite{DBLP:conf/nips/ChenLDZZCDWQLZ24}.

As shown in~\Cref{llava}, vision encoders pretrained with ITO consistently outperform their CLIP counterparts, with the largest improvements on VQAv2, GQA, MMVet, and POPE, and pretraining on DataComp-1B gives the highest overall performance.

\subsection{Ablation Study} 
\label{sec:analysis}
We conduct ablation experiments on YFCC15M and DataComp-1B using ViT-B/16 to analyze the contribution of each component in ITO.

\begin{table}[t]
\centering
\caption{Decomposing the contributions of multi-view alignment and fusion regularization on YFCC15M and DataComp-1B (ViT-B/16, $1$ epoch on $1$B). $\Delta$ values are gains over the previous row. Multi-view alignment provides the dominant gain at both scales; fusion regularization adds a smaller but consistent improvement that grows at billion scale.}
\label{tab:ablation_decomp}
\setlength{\tabcolsep}{5pt}
\renewcommand{\arraystretch}{1.1}
\small
\begin{tabular}{llcccc}
\toprule
Dataset & Setting & ZS IN-1k & Linear IN-1k & MSCOCO I$\to$T & MSCOCO T$\to$I \\
\midrule
\multirow{3}{*}{$15$M}
  & OpenCLIP                          & 36.4 & 51.8 & 26.4 & 15.1 \\
  & + Multi-view alignment ($\lambda{=}0$) & 43.7 \scriptsize{(+7.3)} & 60.2 \scriptsize{(+8.4)} & 30.6 \scriptsize{(+4.2)} & 19.6 \scriptsize{(+4.5)} \\
  & + Fusion regularization ($\lambda{=}2$) & \textbf{44.3} \scriptsize{(+0.6)} & \textbf{60.5} \scriptsize{(+0.3)} & \textbf{30.8} \scriptsize{(+0.2)} & 19.5 \scriptsize{($-$0.1)} \\
\midrule
\multirow{3}{*}{1B}
  & OpenCLIP                          & 63.1 & 67.4 & 47.1 & 29.5 \\
  & + Multi-view alignment ($\lambda{=}0$) & 64.8 \scriptsize{(+1.7)} & 69.6 \scriptsize{(+2.2)} & 49.0 \scriptsize{(+1.9)} & 30.7 \scriptsize{(+1.2)} \\
  & + Fusion regularization ($\lambda{=}2$) & \textbf{65.9} \scriptsize{(+1.1)} & \textbf{69.9} \scriptsize{(+0.3)} & \textbf{49.3} \scriptsize{(+0.3)} & \textbf{31.3} \scriptsize{(+0.6)} \\
\bottomrule
\end{tabular}
\vspace{-2em}
\end{table}

\textbf{Decomposing the contributions of multi-view alignment and fusion regularization.}
\Cref{tab:ablation_decomp} cleanly separates the two components by progressively adding them to a CLIP baseline at both medium ($15$M) and large ($1$B) data scales. Multi-view alignment alone (ITO with $\lambda{=}0$) accounts for the dominant share of the improvement: at YFCC15M it adds $+7.3$ on zero-shot ImageNet-1K, $+8.4$ on linear probing, and $+4.2/+4.5$ on MSCOCO retrieval; the corresponding gains at DataComp-1B are $+1.7/+2.2/+1.9/+1.2$. On top of this, fusion regularization ($\lambda{=}2$) adds a smaller but consistent gain in all four metrics, and---notably---this gain \emph{grows} with data scale: zero-shot accuracy improves by $+0.6$ at 15M and by $+1.1$ at 1B. The accuracy contribution of fusion is thus modest in absolute terms, and at 15M it slightly lowers MSCOCO T$\to$I ($-0.1$). We therefore do not present fusion as a primary source of accuracy; in~\Cref{align_integration} we characterize it as an optimization regularizer.

\begin{table}
\centering
\caption{Ablation study on YFCC15M. We analyze the effect of Multimodal Multiple Alignment and training-time multimodal fusion by varying the fusion weight $\lambda$ (left) and the number of Sentence Blocks (right). Results are zero-shot Top-1 accuracy (\%) on ImageNet-1K using ViT-B/16. The baseline is standard CLIP. }
\label{tab::ablation}
\begin{minipage}{0.4\linewidth}
\centering
\label{tab::lamda}
\begin{tabular}{lcc}
\toprule
Method & $\lambda$ & Top-1 \\
\midrule
Baseline & - & 36.4 \\
\midrule
ITO & 0 & 43.7 \\
\rowcolor{gray!15}  
ITO & 2 & \textbf{44.3} \\
ITO & 4 & 44.0 \\
ITO & 6 & 43.2 \\
ITO & 8 & 43.6 \\
\bottomrule
\end{tabular}
\end{minipage}
\vspace{-1em}
\hspace{0.02\linewidth}
\begin{minipage}{0.4\linewidth}
\centering
\label{tab::depth}
\begin{tabular}{lcc}
\toprule
Method & Blocks & Top-1 \\
\midrule
Baseline & - & 36.4 \\
\midrule
ITO & 1 & 43.9 \\
\rowcolor{gray!15}  
ITO & 2 & \textbf{44.3} \\
ITO & 3 & 43.7 \\
ITO & 4 & 43.5 \\
ITO & 5 & 44.0 \\
\bottomrule
\end{tabular}
\end{minipage}
\vspace{-1em}
\end{table}
\textbf{Sensitivity to fusion weight and depth.}
We further sweep the fusion loss weight $\lambda$ and the number of fusion blocks (\Cref{tab::ablation}). Accuracy follows an inverted U in $\lambda$: it peaks at $\lambda{=}2$ (44.3), and an over-strong fusion signal falls \emph{below} the no-fusion setting ($\lambda{=}6$: 43.2 vs.\ $\lambda{=}0$: 43.7). This is the signature of a regularizer with a bounded optimum: moderate fusion helps, whereas a dominant fusion signal overrides modality-specific structure the dual encoder still needs. A shallow two-block design suffices: deeper fusion modules yield no additional gain, consistent with fusion serving as a training-time regularizer rather than a deep reasoning module. We further explored cross-attention as an alternative fusion architecture and found it consistently underperforms self-attention (Appendix~\ref{appendix:fusion_arch}); alternative fusion read-outs are compared in Appendix~\ref{app:readout}.

\subsection{Are the Gains from Compute or Augmentation?}
\label{sec:compute}

\begin{table}[t]
\centering
\caption{\textbf{Left:} compute-matched comparison on YFCC15M (ViT-B/16); FLOPs are training FLOPs relative to CLIP at 30 epochs. \textbf{Right:} training time (h) and peak memory (GB/GPU) on 64 A800 GPUs, with ratios to OpenCLIP. Inference cost of ITO is identical to CLIP.}
\label{tab:cost}
\small
\setlength{\tabcolsep}{3pt}
\begin{minipage}[t]{0.56\linewidth}
\centering
\begin{tabular}{lcccc}
\toprule
Model (epochs) & FLOPs & ZS & Lin. & T$\to$I \\
\midrule
CLIP (30)              & 1.0$\times$ & 36.4 & 51.8 & 15.1 \\
CLIP (60)              & 2.0$\times$ & 37.3 & 52.8 & 15.4 \\
ITO, $\lambda{=}0$ (30) & 2.3$\times$ & 43.7 & 60.2 & 19.6 \\
ITO, $\lambda{=}2$ (30) & 2.3$\times$ & 44.3 & 60.5 & 19.5 \\
ITO, $\lambda{=}0$ (60) & 4.6$\times$ & 44.6 & 61.7 & 18.7 \\
ITO, $\lambda{=}2$ (60) & 4.6$\times$ & 45.3 & 62.1 & 19.6 \\
\bottomrule
\end{tabular}
\end{minipage}\hfill
\begin{minipage}[t]{0.42\linewidth}
\centering
\begin{tabular}{llcc}
\toprule
Data & Method & Time & Mem. \\
\midrule
\multirow{2}{*}{15M} & OpenCLIP & 3.6 & 7.9 \\
                     & ITO      & 8.2 \scriptsize{(2.3$\times$)} & 16.8 \scriptsize{(2.1$\times$)} \\
\midrule
\multirow{2}{*}{1B}  & OpenCLIP & 8.1 & 30.0 \\
                     & ITO      & ${\sim}18$ \scriptsize{(2.2$\times$)} & 68.6 \scriptsize{(2.3$\times$)} \\
\bottomrule
\end{tabular}
\end{minipage}
\vspace{-1em}
\end{table}

Multi-view alignment processes two image views per caption, so ITO costs about $2.2\times$ the training time of OpenCLIP (\Cref{tab:cost}, right); the fusion regularizer adds negligible cost, and inference cost is unchanged. We test whether the gains come from this extra compute or from the additional augmentation.

\textbf{Compute.} Doubling CLIP's schedule from 30 to 60 epochs improves zero-shot ImageNet-1K by only $+0.9$ (\Cref{tab:cost}, left). CLIP at 60 epochs already uses about 87\% of the FLOPs of ITO at 30 epochs, yet remains $7.0$ points behind. When both are trained for 60 epochs, the gap widens slightly (45.3 vs.\ 37.3), so the advantage of ITO is not erased by a longer schedule.

\textbf{Augmentation.} SLIP~\cite{SLIP} also uses multiple image views but keeps one-to-one cross-modal supervision. On the 26-benchmark zero-shot average (\Cref{tab::classification}), SLIP improves over CLIP (21.4 vs.\ 19.3 on CC3M; 36.0 vs.\ 34.1 on CC12M), and ITO exceeds SLIP by a further $+3.5$ and $+2.4$ points. As SLIP also trains a self-supervised image branch, it is a conservative one-to-one reference; the residual margin is attributable to the many-to-one restructuring rather than to more views.

\subsection{What Does Training-Time Fusion Do?}
\label{align_integration}

\textbf{Fusion as an optimization regularizer.}
Three observations, each comparing ITO with $\lambda{=}2$ against the alignment-only variant ($\lambda{=}0$), which differ only in the fusion module, support this interpretation. \emph{(i) Its gain grows with data scale and training length:} zero-shot ImageNet-1K improves by $+0.6$ at 15M and $+1.1$ at 1B, and MSCOCO T$\to$I changes from $-0.1$ to $+0.6$ (\Cref{tab:ablation_decomp}); on YFCC15M, extending training from 30 to 60 epochs raises the T$\to$I gain from $-0.1$ to $+0.9$ (\Cref{tab:cost}). \emph{(ii) It suppresses late-stage overfitting:} on YFCC15M, CLIP and SLIP degrade in later epochs, and alignment alone raises accuracy without preventing this decline, whereas adding fusion holds zero-shot accuracy flat through the end of training (Appendix~\Cref{fig:training_curve}). \emph{(iii) Its effect is non-monotone in $\lambda$:} an over-strong fusion signal falls below the no-fusion setting (\Cref{sec:analysis}), as expected of a regularizer rather than of an additional source of supervision.

\textbf{Effect on representation geometry.}
Fusion improves the cross-modal consistency score of CyCLIP~\cite{cyclip}, which measures the agreement between top-$K$ neighbors retrieved in image and in text space, at every $K$ and at both scales (\eg $+0.6$ and $+0.8$ at top-1 on YFCC15M and DataComp-1B; Appendix~\Cref{tab:consistency}). Global quantities, however, tell a less uniform story (Appendix~\ref{app:geometry}): fusion narrows the image--text centroid gap~\cite{liang2022mind} at CC3M (centroid cosine $0.423{\to}0.466$) but widens it at DataComp-1B ($0.336{\to}0.314$), while cross-modal MMD and the same-/different-class distance ratio change only at the third to fourth decimal. No single global quantity tracks the downstream gains, consistent with the complex relation between the modality gap and downstream performance reported in~\cite{liang2022mind}. We therefore do not claim that fusion works by improving global geometry, and rest its justification on the optimization evidence above.

\section{Conclusion}
\label{sec:con}
We proposed \textbf{ITO}, whose two components play distinct roles. Multi-view alignment restructures cross-modal supervision from one-to-one to many-to-one and provides the dominant share of discriminative gain, which is not explained by extra compute or augmentation. A training-only fusion module acts as an optimization regularizer whose contribution grows with data scale and which suppresses late-stage overfitting. Because the fusion module is discarded at inference, ITO keeps the cost of a standard dual encoder, suiting inference-cost-constrained settings such as large-scale retrieval and embedding serving.

\textbf{Limitations.}
ITO requires about $2.2\times$ the training time and $2.1$--$2.3\times$ the peak memory of CLIP, almost entirely due to multi-view alignment. At 100M--1B scale we compare mainly against CLIP, so we do not establish superiority over stronger baselines there. The accuracy gain from fusion is modest, and its effect on global geometry is small and scale-dependent; the consistency score is an indirect measure. Results may depend on the augmentation policy and text-view construction (as \name\_sub2 shows). Gains hold across corpora of varying curation (Appendix~\ref{app:dataquality}), but we have not run a controlled noisy-versus-curated ablation. Finally, an over-strong fusion signal hurts accuracy, so reducing modality-induced separation can conflict with preserving modality-specific information.

\begin{ack}
  This work is supported by National Natural Science Foundation of China (U22B2017), and International Cooperation Foundation of Hubei Province, China (2024EHA032).
\end{ack}

{
\small
\bibliographystyle{bib_style}
\bibliography{reference}
}

\newpage
\appendix

\section{Datasets}
\begin{table}[h!]
\caption{The original and actual downloaded number of image-text pairs in pretraining datasets.}
\label{datasets}
\centering\resizebox{0.95\linewidth}{!}{
\begin{tblr}{
  hline{1,4} = {1-6}{1pt},
  hline{2} = {-}{},
  column{even} = {c},
  column{3} = {c},
  column{5} = {c},
}
Dataset       & CC3M~\cite{cc3m}      & CC12M~\cite{cc12m}    & YFCC15M~\cite{orralbaE11} 
& Laion100M~\cite{laion400m} & DataComp-1B~\cite{datacomp-1b}   \\
Origin   & 3.32M & 12.42M  & 15.39M
& 361.02M & 1.40B \\
Download & 2.91M & 10.97M & 14.08M 
& 107.75M & 1.03B 
\end{tblr}
}
\end{table}
We conduct pretraining experiments on several widely used image--text datasets spanning different scales.
All web-scale datasets, including CC3M~\cite{cc3m}, CC12M~\cite{cc12m}, Laion100M (a 100M subset of LAION-400M~\cite{laion400m}), and DataComp-1B~\cite{datacomp-1b}, are downloaded following the official \texttt{img2dataset} pipeline\footnote{\url{https://github.com/rom1504/img2dataset}} with default configurations.
For YFCC15M, we use the filtered English subset released by~\cite{clip}, which can be obtained via the public downloader\footnote{\url{https://github.com/AdamRain/YFCC15M_downloader.git}}.

Due to network availability and filtering, the number of successfully retrieved image--text pairs may differ from the nominal dataset size.
The exact number of samples used for each dataset is summarized in Table~\ref{datasets}.

\section{Implementation Details}
\begin{table}[t!]
\centering
\caption{Common ITO hyperparameters on Datacomp-1B.}
\label{tab::settings}
\begin{tabular}{l|c}
\toprule
Config & Value \\
\midrule
Optimizer & AdamW  \\
Learning rate  & 5e-4 \\
Weight decay & 0.1 \\
Optimizer momentum & $\beta_1$=0.9, $\beta_2$=0.98\\
Batch size & 16384(ViT-B), 8192(ViT-L) \\
Warm-up iterations & 500 \\
$\lambda$ & 2 \\
Scale & (0.5,1.0) \\
Gray\_scale\_prob& 0.2 \\
Color\_jitter& [0.4, 0.4, 0.4, 0.1] \\ Color\_jitter\_prob&0.8\\
\bottomrule
\end{tabular}
\end{table}

\Cref{tab::settings} summarizes the common hyperparameter settings used for
pretraining ITO on the billion-scale DataComp-1B dataset.
Unless otherwise specified, these configurations are shared across all ITO variants
and baselines to ensure fair comparison.

All models are trained using the OpenCLIP framework.
We adopt AdamW as the optimizer with standard momentum parameters and apply cosine
learning-rate decay with linear warmup.
The fusion loss weight $\lambda$ is set to its default value as determined in the
ablation study, and the fusion module consists of two lightweight Transformer blocks.
Image resolution, tokenizer settings, batch size, and training schedules follow the
standard CLIP protocol.

For baselines, we strictly follow the officially recommended hyperparameter settings.
When reproducing baseline results, we use the same dataset versions and training
pipelines as those used for ITO.

\subsection{Image Augmentation Policy}
\label{app:aug}
Each image is augmented independently twice with the same stochastic policy to produce the two views used in multi-view alignment. The policy applies random resized cropping to $224{\times}224$ with scale range $(0.5, 1.0)$, color jitter (brightness $0.4$, contrast $0.4$, saturation $0.4$, hue $0.1$) with probability $0.8$, and random grayscale conversion with probability $0.2$, followed by normalization with the CLIP mean and standard deviation. The CLIP and SLIP baselines use their official augmentation settings.

\section{Fusion Architecture: Self- vs Cross-Attention}
\label{appendix:fusion_arch}

A natural alternative to our self-attention fusion module is cross-attention, where one modality's tokens attend to the other. We compare both designs by replacing the two-layer self-attention block with a two-layer cross-attention block that uses the text EOS token as the query, keeping all other settings unchanged. \Cref{tab:fusion_arch} shows that cross-attention consistently underperforms self-attention; on YFCC15M, it actually drops below the no-fusion baseline ($\lambda{=}0$) on zero-shot accuracy.

We attribute this to the asymmetric, compressive nature of single-query cross-attention: routing all cross-modal information through one query token weakens fine-grained bidirectional token mixing. Self-attention over concatenated tokens, in contrast, lets all multimodal tokens interact symmetrically, which is better suited to lightweight structural regularization where no single token should dominate the bottleneck. We therefore adopt self-attention as the default fusion architecture in ITO.

\begin{table}[h]
\centering
\caption{Self-attention vs cross-attention fusion (two-layer, $\lambda{=}2$). Cross-attention with an EOS-query design consistently underperforms self-attention and, on YFCC15M, falls below the no-fusion variant. Best results are in \textbf{bold}.}
\label{tab:fusion_arch}
\setlength{\tabcolsep}{6pt}
\renewcommand{\arraystretch}{1.05}
\small
\begin{tabular}{llcccc}
\toprule
Dataset & Fusion & ZS IN-1k & Linear IN-1k & MSCOCO I$\to$T & MSCOCO T$\to$I \\
\midrule
\multirow{3}{*}{CC3M}
  & no fusion ($\lambda{=}0$)       & 23.0          & ---           & ---           & --- \\
  & cross-attention                  & 23.0          & 47.5          & 19.7          & 13.6 \\
  & \textbf{self-attention (ours)}   & \textbf{23.3} & \textbf{48.1} & \textbf{21.6} & \textbf{14.6} \\
\midrule
\multirow{3}{*}{YFCC15M}
  & no fusion ($\lambda{=}0$)       & 43.7          & 60.2          & 30.6          & 19.6 \\
  & cross-attention                  & 43.1          & 59.9          & 30.8          & 19.4 \\
  & \textbf{self-attention (ours)}   & \textbf{44.3} & \textbf{60.5} & \textbf{30.8} & \textbf{19.6} \\
\bottomrule
\end{tabular}
\end{table}

\section{Fusion Read-out Ablation}
\label{app:readout}
The fusion module reads out the output at the first position of the joint sequence, i.e., the image \texttt{[CLS]} token, as the fused representation (\Cref{sec:fusion}). The submitted version and our rebuttal described this read-out as the end-of-text (EOT) token; the implementation used for all reported results reads out the image \texttt{[CLS]} token, and we correct the description here. No reported number is affected.

We compare this choice with alternative read-outs, changing only the read-out and keeping all other settings unchanged: mean-pooling over all joint tokens, and an additional learnable \texttt{[FUSE]} token. The \texttt{[FUSE]} token is prepended at the first position of the joint sequence, before the image \texttt{[CLS]} token, without positional embedding, and its output is used as the fused representation. These runs use YFCC15M, ViT-B/16, $\lambda{=}2$, and 30 epochs with the same hyperparameters as the main experiments, but were trained on 16 NVIDIA H200 GPUs (global batch size 4096), whereas the main experiments were trained on NVIDIA A800 GPUs. The numbers in \Cref{tab:readout} are therefore comparable with each other but not directly with the main tables.

\begin{table}[h]
\centering
\caption{Fusion read-out ablation on YFCC15M (ViT-B/16, $\lambda{=}2$, 30 epochs, trained on H200 GPUs). Zero-shot ImageNet-1K top-1 accuracy (\%).}
\label{tab:readout}
\small
\begin{tabular}{lc}
\toprule
Read-out & ZS IN-1k \\
\midrule
Image \texttt{[CLS]} token (default) & \textbf{45.06} \\
Mean-pooling over joint tokens       & 43.90 \\
Learnable \texttt{[FUSE]} token      & \textbf{45.06} \\
\bottomrule
\end{tabular}
\end{table}

A dedicated learnable token performs on par with the image \texttt{[CLS]} token, indicating that the choice between single-token read-outs is not load-bearing. Mean-pooling is $1.2$ points lower. We attribute this to the imbalance of the joint sequence: the 196 image patch tokens outnumber the text tokens, so an unweighted average is dominated by visual content and dilutes the cross-modal signal that a single attending token can aggregate.
\section{Zero-shot Results on DataComp-1B}
\label{scale_1b}

\begin{table}[t!]
\caption{Top-1 zero-shot classification accuracy on 26 public benchmarks following the EVA-CLIP~\cite{DBLP:journals/corr/abs-2303-15389} protocol. Benchmarks include ImageNet-1K~\cite{deng2009imagenet}, ImageNet-A~\cite{inadv}, ImageNet-R~\cite{inren}, CIFAR-10/100~\cite{cifar}, Food-101~\cite{bossard2014food}, Pets~\cite{parkhi12a}, SUN397~\cite{xiao2010sun}, FGVC Aircraft~\cite{maji2013fine}, EuroSAT~\cite{helber2019eurosat}, VOC2007~\cite{everingham2015pascal}, etc.
The best results are highlighted in \textbf{bold}. }
\label{tab::datacomp_b10}
\centering
\scriptsize
\tablestyle{1.4pt}{1.2}
\resizebox{0.9\linewidth}{!}{%
    \begin{tabular}{l|cccccc|cccccccccccccccccccc|c}
        &
        \rotatebox[origin=l]{90}{\scriptsize{ImageNet-1k}} &
        \rotatebox[origin=l]{90}{\scriptsize{ImageNet-A}} &
        \rotatebox[origin=l]{90}{\scriptsize{ImageNet-R}} &
        \rotatebox[origin=l]{90}{\scriptsize{ImageNet-S}} &
        \rotatebox[origin=l]{90}{\scriptsize{ImageNet-V2}} &
        \rotatebox[origin=l]{90}{\scriptsize{ObjectNet}} &
        \rotatebox[origin=l]{90}{\scriptsize{CIFAR-10}} &
        \rotatebox[origin=l]{90}{\scriptsize{CIFAR-100}} & 
        \rotatebox[origin=l]{90}{\scriptsize{Flowers-102}} & 
        \rotatebox[origin=l]{90}{\scriptsize{Food-101}} & 
        \rotatebox[origin=l]{90}{\scriptsize{Pets}} & 
        \rotatebox[origin=l]{90}{\scriptsize{Stanford Cars}} & 
        \rotatebox[origin=l]{90}{\scriptsize{MNIST}} & 
        \rotatebox[origin=l]{90}{\scriptsize{Caltech}} & 
        \rotatebox[origin=l]{90}{\scriptsize{SUN397}} & 
        \rotatebox[origin=l]{90}{\scriptsize{FGVC Aircraft}} & 
        \rotatebox[origin=l]{90}{\scriptsize{Country-211}} & 
        \rotatebox[origin=l]{90}{\scriptsize{DTD}} & 
        \rotatebox[origin=l]{90}{\scriptsize{EuroSAT}} & 
        \rotatebox[origin=l]{90}{\scriptsize{FER2013}} & 
        \rotatebox[origin=l]{90}{\scriptsize{GTSRB}} & 
        \rotatebox[origin=l]{90}{\scriptsize{PCam}} & 
        \rotatebox[origin=l]{90}{\scriptsize{Rendered SST2}} & 
        \rotatebox[origin=l]{90}{\scriptsize{Resisc45}} & 
        \rotatebox[origin=l]{90}{\scriptsize{STL10}} & 
        \rotatebox[origin=l]{90}{\scriptsize{VOC2007}} &
        \rotatebox[origin=l]{90}{\scriptsize{Avg}}
        \\
        \shline
        
        \multicolumn{27}{c}{\scriptsize (a) \textit{Results on CC3M (ViT-B/16, 30 epochs)}} \\
        \hline
        \scriptsize CLIP & 14.8 & 3.2 & 16.0 & 6.0 & 12.7 & 7.8 & 45.5 & 17.8 & 11.6 & 9.4 & 10.4 & 0.7 & 10.2 & 49.8 & 22.3 & 1.6 & 0.5 & 12.5 & 6.0 & 27.3 & 4.5 & 55.6 & 49.9 & 17.4 & 70.3 & 18.5 & 19.3 \\
        \scriptsize SLIP & 18.3 & 5.1 & 19.6 & 8.1 & 15.5 & 10.5 & 47.0 & 19.0 & 10.9 & 13.0 & 10.4 & 1.2 & 10.1 & 56.6 & 31.2 & 1.6 & 0.7 & 12.6 & 6.4 & 18.1 & 2.8 & 55.9 & 50.1 & 17.5 & 88.1 & 27.2 & 21.4 \\
        \scriptsize SigLIP & 16.6 & 3.6 & 19.6 & 7.6 & 14.6 & 9.8 & 51.3 & 16.8 & 10.6 & 10.1 & 8.0 & 1.0 & 14.0 & 46.2 & 21.1 & 0.9 & 0.6 & 16.4 & 5.6 & 16.4 & 11.0 & 43.2 & 50.1 & 24.4 & 66.6 & 16.6 & 19.3 \\
        \scriptsize FLAIR & 19.3 & 4.6 & 19.4 & 8.5 & 16.6 & 11.3 & 63.2 & 29.6 & 13.5 & 12.8 & 9.9 & 1.0 & 8.2 & 61.6 & 30.6 & 0.9 & 0.7 & 12.5 & 4.7 & 26.8 & 5.2 & 57.2 & 50.1 & 27.6 & 83.1 & 25.2 & 23.2 \\
        \scriptsize \textbf{ITO} & 23.3 & 5.6 & 29.0 & 16.2 & 19.9 & 11.8 & 65.0 & 26.1 & 17.1 & 13.7 & 14.3 & 1.7 & 17.9 & 62.9 & 32.4 & 1.1 & 0.7 & 12.7 & 13.8 & 21.4 & 9.3 & 48.4 & 50.1 & 25.9 & 85.6 & 21.9 & \bf24.9 \\
        \scriptsize \textbf{ITO\_sub2} & 23.1 & 4.7 & 28.1 & 16.2 & 19.8 & 12.2 & 57.7 & 26.9 & 13.6 & 13.6 & 13.7 & 1.7 & 10.1 & 66.1 & 31.5 & 1.0 & 0.8 & 11.1 & 14.9 & 26.9 & 6.8 & 59.6 & 50.0 & 27.6 & 88.1 & 21.4 & \bf24.9 \\
        \hline
        
        \multicolumn{27}{c}{\scriptsize (b) \textit{Results on CC12M (ViT-B/16, 30 epochs)}} \\
        \hline
        \scriptsize CLIP & 36.7 & 8.5 & 45.6 & 24.5 & 31.5 & 22.4 & 70.4 & 38.1 & 32.5 & 45.6 & 59.7 & 21.0 & 9.6 & 72.0 & 42.6 & 2.2 & 4.7 & 17.6 & 5.7 & 29.6 & 12.9 & 54.3 & 50.0 & 36.4 & 88.0 & 24.2 & 34.1 \\
        \scriptsize SLIP & 40.9 & 12.4 & 50.7 & 28.8 & 34.7 & 27.6 & 72.6 & 46.6 & 32.9 & 48.8 & 54.0 & 20.7 & 9.0 & 74.9 & 43.8 & 2.7 & 5.4 & 22.6 & 11.5 & 31.7 & 7.9 & 50.0 & 48.8 & 32.1 & 91.3 & 33.2 & 36.0 \\
        \scriptsize SigLIP & 40.6 & 11.3 & 53.0 & 28.7 & 35.1 & 25.2 & 76.5 & 41.0 & 27.8 & 38.1 & 48.4 & 20.9 & 12.9 & 71.4 & 42.1 & 2.6 & 5.0 & 27.1 & 10.2 & 20.9 & 11.4 & 53.5 & 49.6 & 41.2 & 88.7 & 27.5 & 35.0 \\
        \scriptsize FLAIR & 41.5 & 10.4 & 51.2 & 30.2 & 34.9 & 25.5 & 73.5 & 38.3 & 30.5 & 47.6 & 60.8 & 22.2 & 4.6 & 75.4 & 47.9 & 2.0 & 6.2 & 19.4 & 7.2 & 22.8 & 10.7 & 51.8 & 49.9 & 39.3 & 89.1 & 29.6 & 35.5 \\
        \scriptsize \textbf{ITO} & 45.5 & 12.7 & 60.6 & 36.0 & 39.3 & 28.6 & 76.7 & 47.5 & 34.9 & 54.1 & 62.0 & 30.0 & 10.3 & 79.1 & 50.8 & 3.2 & 5.7 & 23.7 & 1.2 & 19.7 & 7.7 & 50.0 & 50.2 & 44.9 & 95.4 & 27.7 & 38.4 \\
        \scriptsize \textbf{ITO\_sub2} & 47.0 & 13.1 & 60.9 & 38.4 & 40.6 & 29.5 & 84.2 & 51.2 & 39.9 & 57.0 & 69.7 & 33.1 & 11.9 & 77.4 & 53.3 & 3.7 & 7.8 & 25.5 & 4.2 & 17.6 & 13.1 & 50.9 & 49.9 & 45.3 & 94.8 & 30.4 & \bf40.4 \\
        \hline

        \multicolumn{27}{c}{\scriptsize (c) \textit{Results on YFCC15M (ViT-B/16, 30 epochs)}} \\
        \hline
        \scriptsize CLIP & 36.4 & 20.2 & 22.9 & 9.4 & 31.9 & 17.6 & 71.6 & 36.9 & 53.6 & 41.6 & 22.6 & 2.7 & 10.2 & 65.1 & 37.6 & 2.7 & 7.1 & 16.7 & 16.6 & 26.3 & 4.4 & 51.9 & 50.0 & 24.6 & 86.1 & 25.6 & 30.5\\
        \scriptsize \textbf{ITO} & 44.3 & 25.6 & 32.7 & 16.8 & 38.6 & 22.8 & 70.4 & 40.6 & 60.6 & 50.8 & 30.9 & 4.5 & 10.5 & 72.8 & 47.2 & 2.4 & 8.4 & 22.4 & 7.1 & 24.3 & 7.2 & 55.0 & 50.4 & 31.7 & 94.0 & 23.8 & 34.5 \\
        \scriptsize \textbf{ITO\_sub2} & 44.4 & 25.2 & 34.1 & 16.0 & 39.1 & 22.5 & 75.5 & 38.3 & 63.2 & 54 & 36.9 & 3.9 & 10.0 & 74.0 & 48.9 & 3.8 & 8.8 & 26.2 & 7.8 & 22.2 & 8.2 & 53.3 & 50.0 & 29.4 & 95.2 & 28.8 & \bf35.4\\
        \hline

        \multicolumn{27}{c}{\scriptsize (d) \textit{Results on Laion100M (ViT-B/16, 30 epochs)}} \\
        \hline
        \scriptsize CLIP & 59.0 & 21.8 & 69.2 & 44.1 & 51.4 & 43.7 & 91.0 & 69.8 & 54.8 & 79.4 & 82.7 & 76.6 & 63.6 & 81.5 & 61.7 & 11.1 & 13.5 & 45.4 & 3.1 & 36.4 & 38.2 & 51.0 & 54.6 & 56.0 & 95.5 & 34.9 & 53.5\\
        \scriptsize SLIP & 55.4 & 21.0 & 65.7 & 41.7 & 48.5 & 43.1 & 89.9 & 68.9 & 51.5 & 75.3 & 74.3 & 72.7 & 57.0 & 81.2 & 58.6 & 9.2 & 11.5 & 45.3 & 1.8 & 42.9 & 35.5 & 50.3 & 50.6 & 50.7 & 93.8 & 36.2 & 51.3 \\
        \scriptsize \textbf{ITO} & 62.8 & 25.6 & 75.5 & 50.2 & 54.8 & 46.2 & 93.8 & 75.0 & 61.4 & 81.9 & 83.6 & 81.7 & 58.1 & 83.8 & 64.6 & 13.7 & 13.9 & 50.1 & 2.7 & 40.6 & 40.7 & 50.3 & 54.4 & 60.2 & 97.0 & 35.8 & \bf56.1\\
        \hline

        \multicolumn{27}{c}{\scriptsize (e) \textit{Results on DataComp-1B (ViT-B/16, 1 epoch)}} \\
        \hline
        \scriptsize CLIP & 63.1 & 24.1 & 67.9 & 47.3 & 54.7 & 50.0 & 92.1 & 74.6 & 63.7 & 83.1 & 84.0 & 76.9 & 55.1 & 82.1 & 62.8 & 14.2 & 13.9 & 42.1 & 3.1 & 25.0 & 38.8 & 50.8 & 49.8 & 58.1 & 95.3 & 35.4 & 54.2\\
        \scriptsize \textbf{ITO} & 65.9 & 25.0 & 72.9 & 53.3 & 57.2 & 51.9 & 94.9 & 78.3 & 70.5 & 83.9 & 86.6 & 83.5 & 57.3 & 83.4 & 64.9 & 18.1 & 14.7 & 46.0 & 3.9 & 24.4 & 35.5 & 53.5 & 51.7 & 60.2 & 96.4 & 37.8 & 56.6 \\
        \scriptsize \textbf{ITO\_sub2} & 65.7 & 26.9 & 74.0 & 53.5 & 57.0 & 52.2 & 95.1 & 78.5 & 67.0 & 84.5 & 85.8 & 82.3 & 57.0 & 82.0 & 64.7 & 16.3 & 14.9 & 47.8 & 3.2 & 29.5 & 42.1 & 58.4 & 51.4 & 57.8 & 96.7 & 37.4 & \bf57.0 \\
        \hline

        \multicolumn{27}{c}{\scriptsize (f) \textit{Results on DataComp-1B (ViT-B/16, 10 epochs)}} \\
        \hline
        \scriptsize CLIP & 72.0 & 45.7 & 81.2 & 58.6 & 64.7 & 62.7 & 95.7 & 80.8 & 74.4 & 90.2 & 92.0 & 86.7 & 79.3 & 84.7 & 69.7 & 25.4 & 21.4 & 59.4 & 2.2 & 39.9 & 54.1 & 62.8 & 51.3 & 66.1 & 97.7 & 37.3 & 63.7 \\
        \scriptsize \textbf{ITO} & 73.5 & 45.7 & 83.2 & 62.0 & 66.3 & 63.6 & 96.9 & 83.5 & 75.8 & 90.8 & 91.1 & 89.2 & 81.1 & 85.1 & 70.3 & 30.8 & 21.8 & 61.4 & 7.3 & 37.8 & 55.5 & 57.7 & 50.5 & 69.9 & 98.3 & 38.3 & \bf64.9 \\
        \scriptsize \textbf{ITO\_sub2} & 72.1 & 42.8 & 82.9 & 60.7 & 64.4 & 62.9 & 96.7 & 82.7 & 74.0 & 89.7 & 90.8 & 87.3 & 76.6 & 85.4 & 69.8 & 27.2 & 21.0 & 60.4 & 4.8 & 32.4 & 45.9 & 56.5 & 52.4 & 66.7 & 98.1 & 38.1 & 63.2\\
        \hline

        \multicolumn{27}{c}{\scriptsize (g) \textit{Results on DataComp-1B (ViT-L/16, 1 epoch)}} \\
        \hline
        \scriptsize CLIP & 69.5 & 39.8 & 78.0 & 56.5 & 61.1 & 60.1 & 96.1 & 80.9 & 71.5 & 88.5 & 88.1 & 83.2 & 70.1 & 83.2 & 68.9 & 16.2 & 19.0 & 51.1 & 2.4 & 33.6 & 52.1 & 59.8 & 51.8 & 63.1 & 97.5 & 37.2 & 60.7\\
        \scriptsize \textbf{ITO} & 72.2 & 44.5 & 82.9 & 62.0 & 64.3 & 64.1 & 97.6 & 83.8 & 72.0 & 89.8 & 90.8 & 87.2 & 74.8 & 84.6 & 70.7 & 27.4 & 20.3 & 56.0 & 2.2 & 31.4 & 49.0 & 50.4 & 53.7 & 67.9 & 98.6 & 38.9 & \bf63.0\\
        \scriptsize \textbf{ITO\_sub2} & 70.7 & 38.8 & 80.4 & 59.4 & 62.8 & 60.3 & 97.0 & 82.4 & 73.4 & 87.9 & 90.7 & 86.8 & 66.7 & 83.4 & 69.6 & 21.5 & 19.1 & 53.9 & 3.4 & 33.7 & 47.3 & 52.8 & 52.0 & 67.2 & 98.2 & 38.3 & 61.5\\
        \hline
        \end{tabular}
}
\end{table}
We provide additional zero-shot classification results on the billion-scale
DataComp-1B dataset to analyze the effect of training duration and model scale.
All evaluations follow the same protocol as in the main paper and use standard
CLIP-style prompts.

\paragraph{Effect of Training Epochs.}
We first study the impact of training duration using ViT-B/16.
In addition to the 1-epoch setting reported in the main paper, we further train
models for 10 epochs to examine scaling behavior with respect to optimization.
As shown in ~\Cref{tab::datacomp_b10}, extending training consistently improves
zero-shot performance across most benchmarks.
Compared with CLIP, ITO exhibits more stable gains as training progresses,
indicating improved utilization of large-scale data.

\paragraph{Effect of Model Scale.}
We further evaluate a larger ViT-L/16 model trained for 1 epoch on DataComp-1B.
Despite the shorter training schedule, the ViT-L/16 variant achieves competitive
or superior performance compared with ViT-B/16 trained for longer durations.
This suggests that ITO benefits from model scaling and that training-time fusion
remains effective when increasing model capacity.

Overall, these results demonstrate that ITO scales favorably with both training
epochs and model size on billion-scale image--text data, while preserving the
standard dual-encoder inference architecture.

\section{Robustness Across Data-Quality Regimes}
\label{app:dataquality}
Image--text pretraining is sensitive to data curation. We have not run a controlled ablation that varies caption quality on a fixed image set, but our pretraining corpora already span a wide range of curation: CC3M and CC12M are alt-text corpora built with automatic filtering pipelines, stricter for CC3M and relaxed for CC12M; YFCC15M consists of noisy user-provided titles and descriptions; Laion100M is loosely filtered by CLIP similarity; and DataComp-1B is aggressively filtered. ITO improves the 26-benchmark zero-shot average over CLIP in every one of these regimes (\Cref{tab::classification}): $+5.6$ on CC3M, $+4.3$ on CC12M, $+4.0$ on YFCC15M, $+2.6$ on Laion100M, and $+2.4$ on DataComp-1B. This is indirect but broad evidence that the method is not tied to a particular curation level. A controlled study that trains on curated and noisy subsets of the same corpus would isolate this factor more cleanly, and we leave it to future work.

\section{Linear Probing: Full Results }
\label{appendix:linear}
\begin{figure}[t!]
\centering
\includegraphics[width=\linewidth]{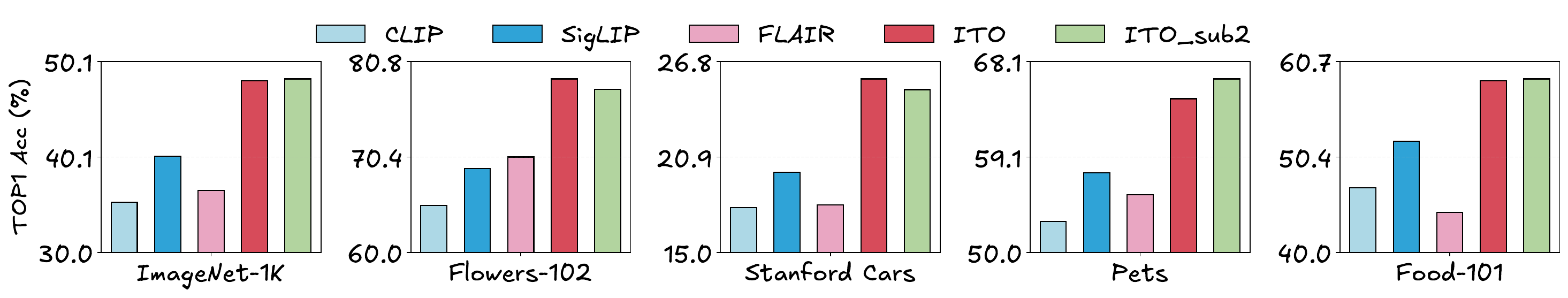}
\caption{
Linear image classification of \name~and its variants pretrained on CC3M.
}
\label{fig:linear_cc3m}
\vspace{-1em}
\end{figure}

\begin{figure}[t!]
\centering
\includegraphics[width=\linewidth]{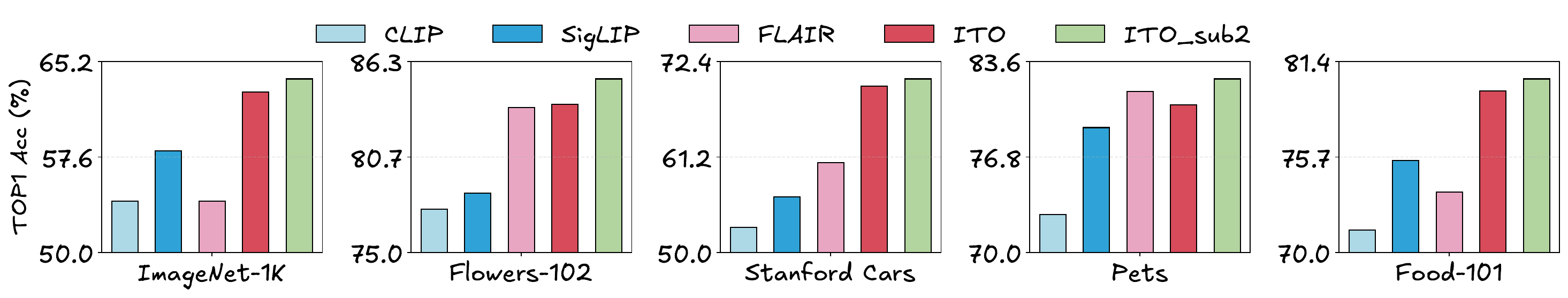}
\caption{
Linear image classification of \name~and its variants pretrained on CC12M.
}
\label{fig:linear_cc12m}
\vspace{-1em}
\end{figure}

This section provides additional linear probing results that complement ~\Cref{tab::linear-recap} in the main paper. We report (i) per-dataset results on five classification benchmarks (ImageNet-1K, Flowers-102, Stanford Cars, Pets, Food-101) across all pretraining scales (~\Cref{linear_full}), and (ii) results on CC3M and CC12M, which are not included in the main-paper table due to space constraints (~\Cref{fig:linear_cc3m} and~\Cref{fig:linear_cc12m}).


\begin{table}[t!]
\caption{Linear image classification of \name~and its variants pretrained on YFCC15M, Laion100M and DataComp-1B. The best results are highlighted in \textbf{bold}. $^\dagger$ indicates models trained for 10 epochs using ViT-B/16. $^*$ indicates models trained for 1 epochs using ViT-L/16.}
\label{linear_full}
\centering
\resizebox{0.85\linewidth}{!}{
\begin{tblr}{
  width = \linewidth,
    column{8} = {c},
    column{3} = {c},
  column{4} = {c},
  column{5} = {c},
  column{6} = {c},
  column{7} = {c},
hline{1,16}={1-8}{1pt},
  cell{2}{1} = {r=3}{},
  cell{5}{1} = {r=2}{},
  cell{7}{1} = {r=9}{},
  hline{2,5,7} = {1-8}{},
  hline{10} = {2-8}{},
  hline{13} = {2-8}{},
  vline{8} = {1-15}{},
}
Dataset  & Method & IN-1k & Flowers & Cars  & Pets  & Food & Avg \\
15M & CLIP & 51.82	& 85.17 & 	19.60 &  53.61 & 69.18 & 55.88\\
 &\textbf{\name} &\bf60.52	& 92.18 & 	\bf28.37 & 61.22 & \bf76.50  & 63.76\\
 & \textbf{\name\_sub2} & 60.40&	\bf92.37	&27.73	&  \bf62.82&	76.22  & \bf63.91\\
100M  & CLIP             & 67.27    & 86.65   & 87.17 & 82.88 & 85.35 & 81.86 \\
          &  \textbf{\name} & \bf68.70     & \bf90.21   & \bf88.16 & \bf84.49 & \bf86.22  & \bf83.56\\
          1B  & CLIP & 67.39 &	90.08 &	87.14&  79.61& 	86.47 & 82.14\\
           & \textbf{\name} & 69.87	& \bf92.47	&\bf90.70	&81.88&	87.75 & 82.53 \\
           & \textbf{\name\_sub2} & \bf70.76&	92.24	&90.34	&\bf83.16	&\bf88.51 & \bf83.00 \\
           &CLIP$^\dagger$ & 74.99&	94.67&	91.64	&87.90& 	91.89 & 87.62\\
          & \textbf{\name}$^\dagger$ & \bf76.20	&\bf96.16	&\bf92.89	&88.09&	\bf92.54 & \bf89.18\\
           & \textbf{\name\_sub2}$^\dagger$  & 75.83&	94.96	&92.23	&\bf89.37	&92.19 & 88.92\\
           & CLIP$^*$ & 73.78&	93.04&	90.23&	86.54&	90.83 & 86.88\\
          & \textbf{\name}$^*$ & \bf75.97&	\bf95.80	&92.28&	87.84	&\bf92.27 & \bf88.83\\
           & \textbf{\name\_sub2}$^*$ & 74.88&	94.89	&\bf92.29	&\bf88.93 &	91.43  & 88.48\\
\end{tblr}
}
\end{table}

\section{Sub-description Protocol}
\label{app:sub2}
Sub-descriptions are generated and sampled following FLAIR~\cite{xiao2024flair}; at each training step, two sub-descriptions are sampled per caption to form the two text views of \name\_sub2.

\section{Additional Experiments on CC3M-recap}
\label{appendix:cc3mrecap}

\begin{table*}[t!]
\caption{Top-1 zero-shot classification accuracy on 26 public benchmarks following the EVA-CLIP~\cite{DBLP:journals/corr/abs-2303-15389} protocol. Results are reported for models pretrained on CC3M-recap with both ViT-B/16 as vision encoders. The best results are highlighted in \textbf{bold}. }
\label{tab::cc3m-recap-cls}
\centering
\scriptsize
\tablestyle{1.4pt}{1.2}
    \resizebox{\textwidth}{!}{\begin{tabular}{l|cccccc|cccccccccccccccccccc|c}
        &
        \rotatebox[origin=l]{90}{\scriptsize{ImageNet-1k}} &
        \rotatebox[origin=l]{90}{\scriptsize{ImageNet-A}} &
        \rotatebox[origin=l]{90}{\scriptsize{ImageNet-R}} &
        \rotatebox[origin=l]{90}{\scriptsize{ImageNet-S}} &
        \rotatebox[origin=l]{90}{\scriptsize{ImageNet-V2}} &
        \rotatebox[origin=l]{90}{\scriptsize{ObjectNet}} &
        \rotatebox[origin=l]{90}{\scriptsize{CIFAR-10}} &
        \rotatebox[origin=l]{90}{\scriptsize{CIFAR-100}} & 
        \rotatebox[origin=l]{90}{\scriptsize{Flowers-102}} & 
        \rotatebox[origin=l]{90}{\scriptsize{Food-101}} & 
        \rotatebox[origin=l]{90}{\scriptsize{Pets}} & 
        \rotatebox[origin=l]{90}{\scriptsize{Stanford Cars}} & 
        \rotatebox[origin=l]{90}{\scriptsize{MNIST}} & 
        \rotatebox[origin=l]{90}{\scriptsize{Caltech}} & 
        \rotatebox[origin=l]{90}{\scriptsize{SUN397}} & 
        \rotatebox[origin=l]{90}{\scriptsize{FGVC Aircraft}} & 
        \rotatebox[origin=l]{90}{\scriptsize{Country-211}} & 
        \rotatebox[origin=l]{90}{\scriptsize{DTD}} & 
        \rotatebox[origin=l]{90}{\scriptsize{EuroSAT}} & 
        \rotatebox[origin=l]{90}{\scriptsize{FER2013}} & 
        \rotatebox[origin=l]{90}{\scriptsize{GTSRB}} & 
        \rotatebox[origin=l]{90}{\scriptsize{PCam}} & 
        \rotatebox[origin=l]{90}{\scriptsize{Rendered SST2}} & 
        \rotatebox[origin=l]{90}{\scriptsize{Resisc45}} & 
        \rotatebox[origin=l]{90}{\scriptsize{STL10}} & 
        \rotatebox[origin=l]{90}{\scriptsize{VOC2007}} &
        \rotatebox[origin=l]{90}{\scriptsize{Avg}}
        \\
        \shline
        
        \multicolumn{27}{c}{\scriptsize (a) CC3M Pre-training ViT-B} \\
        \hline
        \scriptsize FLAIR  &27.7	&7.6	&31.4	&15.0	&23.8	&15.5	&73.9	&44.3	&18.9	&19.8	&25.9	&2.5	&13.1	&69.8	&42.3	&1.6	&2.6	&18.8	&7.4	&30.6	&11.4	&45.7	&50.1	&34.6	&92.9	&26.6 & 29.0\\
        \scriptsize \textbf{ITO\_sub2}     &30.4	&8.8	&38.8	&19.9	&27.2	&16.1	&62.7	&38.0	&19.7	&23.1	&29.9	&3.6	&18.5	&71.4	&46.0	&2.6	&2.6	&20.5	&7.1	&22.4	&11.9	&50.0	&50.1	&37.3	&94.4	&27.1  &  \bf30.0\\
        \hline
        \scriptsize \textbf{ITO\_sub3}      &30.9	&9.4	&39.5	&20.7	&26.7	&16.3	&62.2	&36.1	&20.0	&23.1	&30.2	&4.2	&16.8	&72.1	&46.0	&1.2	&2.6	&21.3	&8.5	&19.6	&10.2	&50.0	&50.1	&35.0	&94.0	&22.5   &29.6  \\
        \hline
    
        \end{tabular}
        }
\end{table*}
To further investigate the impact of text augmentation, we conduct supplementary experiments on the CC3M-recap dataset, a synthetic dataset provided by Dreamlip~\cite{dreamlip}. 
In CC3M-recap, each original image–text pair $(I, T)$ is processed by Vision–Language Models (VLMs) using two prompts—``Describe the image in detail'' and ``Describe the image in short''—to generate a long and a short caption, respectively. 
Moreover, by employing three different VLMs (InstructBLIP~\cite{InstructBLIP}, LLaVA-1.5~\cite{llava15}, and ShareGPT4V~\cite{sharegpt4v}), each original pair yields six additional synthetic textual descriptions. 
This setup allows us to examine whether richer textual supervision can further enhance the performance of the proposed multimodal fusion framework.

\paragraph{Experimental Setup.}
Following the same training configurations as in the main experiments, we compare FLAIR~\cite{xiao2024flair} with our text-augmented variants \name\_sub2 and \name\_sub3 on the CC3M-recap dataset. The original CC3M-recap dataset contains approximately 2.8 million image–text pairs, while we use about 1.7 million successfully downloaded samples in our experiments. For \name\_sub2 and \name\_sub3, we randomly sample two or three sub-descriptions from the long captions to construct multiple text views, respectively. 
All models are trained for 30 epochs using ViT-B/16 as the vision encoder. 
We evaluate them on the same downstream tasks as in the main paper, including 26 zero-shot classification datasets, linear probing for image classification, and image–text retrieval benchmarks.

\paragraph{Results.}
The results are summarized in~\Cref{tab::cc3m-recap-cls,tab::cc3m-recap-linear,tab::cc3m-recap-retrieval,tab::cc3m-recap-fine-grained}. 
Across all downstream tasks, incorporating richer textual supervision from CC3M-recap leads to consistent improvements over FLAIR~\cite{xiao2024flair}. 

In the 26-dataset zero-shot classification benchmark (~\Cref{tab::cc3m-recap-cls}), both \name\_sub2 and \name\_sub3 outperform FLAIR by a clear margin, achieving average Top-1 accuracies of 30.0\% and 29.6\%, respectively. 
This demonstrates that our contrastive–fusion framework benefits from more diverse textual views even when trained on synthetic captions. 

For linear probing (~\Cref{tab::cc3m-recap-linear}), \name\_sub2 and \name\_sub3 again show notable gains on ImageNet-1K and several fine-grained datasets, with \name\_sub3 achieving the best overall results (e.g., +7.4 points over FLAIR on ImageNet-1K). 
These findings indicate that long-caption sampling not only enhances zero-shot transfer but also improves the linear separability of learned visual features.


\begin{table}[t!]
\caption{Linear image classification of \name~and its variants pretrained on CC3M-recap. }
\label{tab::cc3m-recap-linear}
\centering
\resizebox{0.7\linewidth}{!}{
\begin{tblr}{
  width = \linewidth,
    column{3} = {c},
  column{4} = {c},
hline{1, 5}={1-12}{1pt},
  cell{2}{1} = {r=3}{},
  hline{2} = {-}{},
}
Dataset    & Method                       & ImageNet-1k & Flowers-102 & Stanford Cars  & Pets  & Food-101  \\
CC3M-recap & FLAIR   & 45.65    & \bf76.57   & 25.05 & \bf67.05 & 59.38 \\
           & \textbf{\name\_sub2}           & 52.43    & 73.64   & 23.96 & 65.09 & 64.15 \\
           & \textbf{\name\_sub3}          & \bf53.05    & 74.30    & \bf25.32 & 65.55 & \bf64.70  
\end{tblr}
}
\end{table}

\begin{table}[t!]
\caption{Zero-shot image-text retrieval on validation splits for standard benchmarks (MSCOCO and Flickr30k).}
\label{tab::cc3m-recap-retrieval}
\centering
\resizebox{0.7\linewidth}{!}{
\begin{tblr}{
  width = \linewidth,
  hline{1,7}={1-14}{1pt},
  row{2} = {c},
  cell{1}{1} = {r=3}{},
  cell{1}{2} = {r=3}{},
  cell{1}{3} = {c=6}{c}{},
  cell{1}{9} = {c=6}{c}{},
  cell{2}{3} = {c=3}{},
  hline{3}={3-5}{leftpos = -1, rightpos = -1, endpos},
  cell{2}{6} = {c=3}{},
  hline{3}={6-8}{leftpos = -1, rightpos = -1, endpos},
  cell{2}{9} = {c=3}{},
   hline{3}={9-11}{leftpos = -1, rightpos = -1, endpos},
  cell{2}{12} = {c=3}{},
   hline{3}={12-14}{leftpos = -1, rightpos = -1, endpos},
  cell{4}{1} = {r=3}{},
  vline{4} = {1}{},
  vline{7} = {2}{},
  vline{9} = {1-18}{},
  hline{4} = {-}{},
  hline{2} = {3-14}{},
}
Dataset & Method   & MSCOCO &       &       &       &       &       & Flickr30k &       &       &       &       &       \\
        &          & I $\to$ T    &       &       & T $\to$ I   &       &       & I $\to$ T       &       &       & T $\to$ I   &       &       \\
        &          & R@1    & R@5   & R@10  & R@1   & R@5   & R@10  & R@1       & R@5   & R@10  & R@1   & R@5   & R@10  \\
        CC3M-recap & FLAIR  & 37.54     & 64.38 & 75.88     & 29.76 & 55.53   & 66.69    & 65.19     & 87.08 & 92.41    & 53.25     & 78.01 & 85.48    \\
           & \textbf{\name\_sub2}           & \bf47.76    & 73.76    & 82.64   & 31.59     & 58.54   & \bf69.98   & 74.56    & 93.89    & 96.75   & 57.32   & 82.23   & 88.38    \\
           & \textbf{\name\_sub3}          & 47.40     & \bf74.94     & \bf83.86  & \bf32.38 & \bf58.74    &  69.77 & \bf74.95   & \bf94.38    & \bf96.75   & \bf59.13    & \bf83.41     & \bf89.80  \\
        
\end{tblr}
}
\vspace{-1em}
\end{table}

\begin{table}[t!]
\caption{Zero-shot image-text retrieval on validation splits for fine-grained  retrieval benchmarks (DOCCI).}
\label{tab::cc3m-recap-fine-grained}
\centering
\resizebox{0.65\linewidth}{!}{
\begin{tblr}{
  width = \linewidth,
  hline{1,7}={1-14}{1pt},
  row{2} = {c},
  cell{1}{1} = {r=3}{},
  cell{1}{2} = {r=3}{},
  cell{1}{3} = {c=6}{c}{},
  cell{2}{3} = {c=3}{},
  hline{3}={3-5}{leftpos = -1, rightpos = -1, endpos},
  cell{2}{6} = {c=3}{},
  hline{3}={6-8}{leftpos = -1, rightpos = -1, endpos},
  cell{4}{1} = {r=3}{},
  vline{4} = {1}{},
  vline{7} = {2}{},
  vline{6} = {2-18}{},
  hline{4} = {-}{},
  hline{2} = {3-14}{},
}
Dataset & Method & DOCCI &       &       &       &       &       \\
        &        & I $\to$ T    &       &       & T $\to$ I   &       &       \\
        &        & R@1   & R@5   & R@10  & R@1   & R@5   & R@10  \\
        CC3M-recap & FLAIR  & 20.16   & 43.56     & 54.66  &  10.65 & 24.23     & 31.60  \\
           & \textbf{\name\_sub2}           & \bf27.82    & \bf54.14     & \bf65.90    & \bf10.70     & 23.69    & 30.84    \\
           & \textbf{\name\_sub3}           & 27.54    & 53.68    & 65.30     & 10.68     &  \bf24.04 & \bf31.27       
\end{tblr}
}
\end{table}

In zero-shot image–text retrieval (~\Cref{tab::cc3m-recap-retrieval} and~\Cref{tab::cc3m-recap-fine-grained}), our models consistently surpass FLAIR on both coarse-grained (MSCOCO, Flickr30k) and fine-grained (DOCCI) benchmarks. 
For instance, \name\_sub3 improves Recall@1 on MSCOCO I → T from $37.54\%$ to $47.40\%$, and on Flickr30k from $65.19\% $to $74.95\%$. 
Similar gains are observed on DOCCI, confirming that text augmentation yields more robust and fine-grained alignment between images and captions.

Overall, these results validate the effectiveness of our text sampling strategy under the VLM-recaptioned setting. 
Richer and more diverse textual descriptions help our method capture finer semantic correspondences, while maintaining strong generalization across classification and retrieval tasks.

\paragraph{Discussion.}
These findings suggest that when captions are linguistically rich, sub-description sampling further enhances cross-modal learning by emphasizing diverse semantic aspects of the same image. 
Compared with the models pretrained on the original CC3M dataset, those trained on CC3M-recap achieve consistently higher performance across classification and retrieval benchmarks, confirming that richer textual supervision facilitates more effective image–text alignment. 
However, as shown in~\Cref{sec:zs}, this advantage attenuates on DataComp-1B, where native captions are already lexically diverse and sub-descriptions become largely redundant.

\section{DOCCI and Full Zero-shot Retrieval Results}
\label{app:docci}

We provide additional zero-shot image--text retrieval results to complement the
main paper. This section reports the complete Recall@1/5/10 metrics on MSCOCO
and Flickr30k, as well as fine-grained retrieval performance on DOCCI.

\begin{table}[t!]
\caption{Zero-shot image-text retrieval on validation splits for standard benchmarks (MSCOCO and Flickr30k). The best results are highlighted in \textbf{bold}. $^\dagger$ indicates models trained for 10 epochs using ViT-B/16. $^*$ indicates models trained for 1 epochs using ViT-L/16.}
\label{tab::retrieval-full}
\centering
\resizebox{0.9\linewidth}{!}{
\begin{tblr}{
  width = \linewidth,
  hline{1, 28}={1-14}{1pt},
  row{2} = {c},
  cell{1}{1} = {r=3}{},
  cell{1}{2} = {r=3}{},
  cell{1}{3} = {c=6}{c}{},
  cell{1}{9} = {c=6}{c}{},
  cell{2}{3} = {c=3}{},
  hline{3}={3-5}{leftpos = -1, rightpos = -1, endpos},
  cell{2}{6} = {c=3}{},
  hline{3}={6-8}{leftpos = -1, rightpos = -1, endpos},
  cell{2}{9} = {c=3}{},
   hline{3}={9-11}{leftpos = -1, rightpos = -1, endpos},
  cell{2}{12} = {c=3}{},
   hline{3}={12-14}{leftpos = -1, rightpos = -1, endpos},
  cell{4}{1} = {r=5}{},
  cell{9}{1} = {r=5}{},
  cell{14}{1} = {r=3}{},
  cell{17}{1} = {r=2}{},
  cell{19}{1}= {r=9}{},
  vline{5} = {1}{},
  vline{8} = {2}{},
  vline{9} = {1-27}{},
  hline{4,9,14,17,19} = {-}{},
  hline{22} = {2-14}{},
  hline{25} = {2-14}{},
  hline{2} = {3-14}{},
}
Dataset   & Method   & MSCOCO &       &       &       &       &       & Flickr30k &       &       &       &       &       \\
           &       & I $\to$ T    &       &       & T $\to$ I   &       &       & I $\to$ T       &       &       & T $\to$ I   &       &       \\
        &          & R@1    & R@5   & R@10  & R@1   & R@5   & R@10  & R@1       & R@5   & R@10  & R@1   & R@5   & R@10  \\
3M & CLIP & 12.36  & 30.98 & 41.76 & 8.14  & 22.68 & 31.98 & 23.57     & 50.69 & 62.03 & 16.98 & 37.65 & 48.68 \\
        & SigLIP   & 13.70  & 32.66 & 43.16 & 9.28  & 23.48 & 32.60 & 28.90     & 53.25 & 65.09 & 18.32 & 39.43 & 49.45 \\
        & FLAIR    & 17.86  & 38.90 & 51.18 & 12.59 & 30.38 & 40.94 & 35.40     & 62.13 & 74.46 & 25.74 & 49.86 & 60.81 \\
      & \textbf{\name}    & \bf21.56  & 45.36 & 57.22 & \bf14.56 & 33.71 & 44.90 & \bf42.6     & \bf71.2 & \bf80.87 & \bf29.45 & \bf53.89 & \bf64.36 \\
      & \textbf{\name\_sub2 } & 21.08    & \bf45.62     & \bf57.42    & 14.01    & \bf33.88   & \bf45.06    & 42.11    & 69.43     & 78.11 & 28.26   & 53.08    & 64.16    \\
12M   & CLIP & 34.17  & 61.20 & 72.52 & 23.04 & 47.58 & 59.39 & 62.23     & 86.29 & 92.11 & 45.74 & 73.02 & 82.23 \\
        & SigLIP  & 39.98  & 67.28 & 77.82 & 26.85 & 51.48 & 63.20 & 68.34     & 89.45 & 93.29 & 50.87 & 75.96 & 84.18 \\
        & FLAIR    & 36.20  & 63.16 & 74.38 & 24.62 & 48.79 & 60.57 & 62.92     & 88.56 & 93.20 & 47.36 & 74.77 & 83.21 \\
        & \textbf{\name}   & 42.30  & 70.00 & 79.46 & 29.62 & 55.74  & 67.09 & \bf72.49     & 90.83 & 95.07  & 54.50 & 79.98  & 86.86 \\
       & \textbf{\name\_sub2 } & \bf43.94     & \bf70.40    & \bf80.26    & \bf30.51   & \bf56.76   & \bf68.29    & 72.29    & \bf92.80   & \bf96.35     & \bf56.65     & \bf82.54    & \bf88.58    \\
        15M & CLIP & 26.38 &   50.86    &62.86       & 15.06    & 34.82  &  46.43  & 47.14   & 74.46     & 83.23    &   30.77 &   56.82     & 67.14\\
       & \textbf{\name} & 30.76   & 57.44  &  68.82      &  19.57&     41.61    &  53.03 & 54.83   & 82.74  &   88.56     &  35.72 &   62.60    &73.06 \\
       & \textbf{\name\_sub2} & \bf31.42   & \bf58.02   & \bf69.06     &   \bf20.04    & \bf42.88  &   \bf54.61 & \bf56.21  &  \bf83.14    &  \bf90.43  &     \bf37.95 &   \bf65.46  &  \bf75.50\\
100M  & CLIP             & 49.12    & 74.66    & 83.76    & 31.92    & 57.32    & 68.40  & 75.94     & 93.59 & 96.75   & 60.04  & 84.08   & 89.90     \\
           & \textbf{\name} & \bf52.08     &\bf76.04  & \bf84.24 & \bf34.26    & \bf59.57    & \bf70.07 & \bf79.59    & \bf95.66    & \bf98.42   & \bf63.00     & \bf86.21    & \bf91.42    \\
1B  & CLIP  & 47.08   & 72.70  &  82.64      &  29.51&     54.84  &  66.05 & 72.68 &   90.63    &94.48&       54.73&     80.49&    87.53 \\
& \textbf{\name} & 49.26   & \bf74.76&    \bf83.62  &      31.30&     56.63&    68.22 & \bf75.94    & \bf94.08    &\bf97.14  &      \bf56.79   &  82.17&    88.56 \\
& \textbf{\name\_sub2} & \bf49.50 &   74.12  &  83.06      &   \bf31.89  &   \bf57.26    & \bf68.36 & 73.37&    93.00   & 96.15&        55.70   &  \bf82.33 &  \bf89.33\\
& CLIP$^\dagger$ & 57.70    & 80.86 &   87.72  &       37.76   & 63.58  &  73.50& 82.25 &   96.65   & 98.52  &      66.57  &   87.71&    92.60 \\
& \textbf{\name}$^\dagger$ & \bf58.14    &  \bf81.76 &   \bf88.88   &     \bf38.97    & \bf64.45   & \bf74.63 & \bf84.81   &  \bf96.45   & \bf98.52   &     \bf67.10  &   \bf88.38 &   \bf93.18 \\
& \textbf{\name\_sub2}$^\dagger$ & 55.88    & 79.90 &   87.84  &      38.33&    64.03  &  74.03 & 81.95    & 95.36   & 98.03   &     65.88&    87.67    &92.49\\
 & CLIP$^*$ & 53.40   &  77.76    & 86.62   &     34.88   & 60.16  &  71.04 & 78.40   & 95.07   & 97.63  &     62.60    & 85.62   & 91.26   \\
& \textbf{\name }$^*$ & \bf56.58   &  \bf80.46&    \bf88.12     &   \bf38.12   &  \bf63.72  &  \bf73.62 & \bf83.14 &   \bf96.35  &  \bf98.62 &       \bf66.33   &  \bf87.69 &  \bf92.96\\
 & \textbf{\name\_sub2}$^*$ & 52.12 &    77.44    & 85.96   &     35.69  &  61.62 &    71.86 & 78.99  &  95.27   & 98.03      & 62.88  &   85.42  &  91.20 \\
        
\end{tblr}
}
\end{table}
\paragraph{Complete Retrieval Results on MSCOCO and Flickr30k.}
\Cref{tab::retrieval-full} report full bidirectional retrieval results
(Image$\rightarrow$Text and Text$\rightarrow$Image) on the validation splits of
MSCOCO and Flickr30k.
These results extend the main paper, where only Recall@1 and Recall@5 are shown
for clarity.
The trends are consistent with those reported in the main text: ITO achieves
strong and stable improvements over CLIP, SLIP, and FLAIR across datasets and
retrieval directions, and the relative ordering of methods remains unchanged
when considering Recall@10.

\paragraph{Fine-grained Retrieval on DOCCI.}
We further evaluate zero-shot retrieval on the DOCCI dataset, which provides an
average of seven sentence-level descriptions per image and emphasizes fine-grained
cross-modal matching.
As shown in~\Cref{tab::fine-grained}, ITO remains competitive across all recall
metrics, demonstrating that representations learned via training-time fusion
generalize well to detailed and compositional image--text retrieval scenarios.
These results are consistent with the conclusions drawn from COCO and Flickr30k,
and further validate the robustness of ITO under diverse retrieval settings.

\begin{table}[t!]
\caption{Zero-shot image-text retrieval on validation splits for fine-grained  retrieval benchmarks (DOCCI). The best results are highlighted in \textbf{bold}. $^\dagger$ indicates models trained for 10 epochs using ViT-B/16. $^*$ indicates models trained for 1 epochs using ViT-L/16.}
\label{tab::fine-grained}
\centering
\resizebox{0.5\linewidth}{!}{
\begin{tblr}{
  width = \linewidth,
  hline{1, 28}={1-8}{1pt},
  cell{1}{1} = {r=3}{},
  cell{1}{2} = {r=3}{},
  cell{1}{3} = {c=6}{c}{},
  cell{2}{3} = {c=3}{c}{},
  hline{3}={3-5}{leftpos = -1, rightpos = -1, endpos},
  cell{2}{6} = {c=3}{c}{},
  hline{3}={6-8}{leftpos = -1, rightpos = -1, endpos},
  cell{4}{1} = {r=5}{},
  cell{9}{1} = {r=5}{},
  cell{14}{1} = {r=3}{},
   cell{17}{1} = {r=2}{},
   cell{19}{1} = {r=9}{},
  vline{6} = {2-27}{},
  hline{4,9,14,17,19 } = {1-8}{},
  hline{22} = {2-8}{},
  hline{25} = {2-8}{},
  hline{2} = {3-8}{},
}
Dataset& Method & DOCCI &       &       &       &       &       \\
        &        & I $\to$ T    &       &       & T $\to$ I   &       &       \\
       &        & R@1   & R@5   & R@10  & R@1   & R@5   & R@10  \\
3M    & CLIP   & 4.10  & 12.56 & 18.72 & 2.13  & 6.71  & 10.07 \\
        & SigLIP & 4.98  & 14.38 & 20.88 & 2.28  & 6.91  & 10.32 \\
         & FLAIR  & 7.12  & 19.12 & 28.00 & 3.23  & 9.70 & 14.13 \\
         & \textbf{\name}  & 8.24  & 22.00 & \bf31.64 & 3.48   & \bf9.97 & 14.39 \\
          & \textbf{\name\_sub2}  & \bf9.18      & \bf22.74 & 31.20     & \bf3.49    & 9.96    & \bf14.41    \\
12M  & CLIP   & 20.14 & 43.40 & 53.54 & 7.67  & 17.96 & 24.03 \\
      & SigLIP & 25.00 & 49.66 & 61.72 & 9.38  & 20.59 & 26.94 \\
        & FLAIR  & 24.24 & 48.50 & 59.12 & 9.36 & 21.20 & 28.11 \\
        & \textbf{\name}  & 26.62 & 51.00 & 61.88 & 9.39 & 20.51 & 26.92 \\
         & \textbf{\name\_sub2}  & \bf28.26     & \bf53.92  & \bf64.46 & \bf10.86     & \bf23.60  & \bf30.44    \\
15M & CLIP & 15.94 & 35.26 & 45.70 & 5.45 & 13.54 & 18.81 \\
 & \textbf{\name} & 19.06 & 40.28 & 51.90 & 6.82 & 16.45 & 22.45 \\
& \textbf{\name\_sub2} & \bf19.82 & \bf43.06 & \bf53.70 & \bf6.86 & \bf17.52 & \bf23.60\\ 
100M  & CLIP            & 39.28    & 65.30     & 74.68    & 13.04     & 25.45& 31.81    \\
          & \textbf{\name} & \bf41.54     & \bf67.48 & \bf76.52     & \bf13.68 & \bf26.48    & \bf32.80   \\
1B  & CLIP & 43.02 &     69.48  &  77.88   &   13.01   &    25.71   &  32.28 \\
 & \textbf{\name} & 43.24   & 69.98   & 79.26  &   13.75 &      26.78 &    33.27 \\
 &\textbf{\name\_sub2} & \bf44.08 &   \bf70.86  &   \bf79.56  &   \bf14.47    &   \bf28.31  &   \bf35.01 \\
 & CLIP$^\dagger$ & 52.58  &  76.98 &    84.78     & 16.66 &     30.49     & 37.09\\
 & \textbf{\name}$^\dagger$ & \bf52.78   & \bf78.20  &   \bf85.72   &   17.30 &     31.60 &     38.30 \\
& \textbf{\name\_sub2}$^\dagger$  & 50.78  &  76.18    & 83.78   &    \bf17.79   &   \bf32.73   &  \bf39.90 \\
 &  CLIP$^*$ & 48.98 &    75.26 &    83.12  &   15.59 &      29.16   & 35.49      \\
 & \textbf{\name}$^*$ & \bf50.94 &    \bf76.56 &    \bf84.36  &  \bf16.77      & 30.75  &   37.60 \\
 & \textbf{\name\_sub2}$^*$ & 48.34    & 75.02   &  82.68   &  16.67   &   \bf30.95   &  \bf38.01 \\
\end{tblr}
}
\end{table}



\section{Consistency Score and Global Geometry Probes}
\label{app:geometry}

\begin{table}[h]
\centering
\caption{Cross-modal consistency score (\%) at top-$K$ following CyCLIP~\cite{cyclip}. $\Delta$ is the change from adding fusion regularization ($\lambda{=}2$) to the alignment-only variant ($\lambda{=}0$). ZS IN-1k is reported alongside for reference.}
\label{tab:consistency}
\small
\begin{tabular}{llccccc}
\toprule
Dataset & Setting & Top-1 & Top-3 & Top-5 & Top-10 & ZS IN-1k \\
\midrule
\multirow{3}{*}{15M}
  & ITO ($\lambda{=}0$) & 32.1 & 31.9 & 33.7 & 35.3 & 43.7 \\
  & ITO ($\lambda{=}2$) & \textbf{32.7} & \textbf{32.4} & \textbf{34.3} & \textbf{35.7} & \textbf{44.3} \\
  & $\Delta$            & +0.6 & +0.5 & +0.6 & +0.4 & +0.6 \\
\midrule
\multirow{3}{*}{1B}
  & ITO ($\lambda{=}0$) & 46.1 & 45.1 & 47.5 & 49.1 & 64.8 \\
  & ITO ($\lambda{=}2$) & \textbf{46.9} & \textbf{45.7} & \textbf{48.0} & \textbf{49.8} & \textbf{65.9} \\
  & $\Delta$            & +0.8 & +0.6 & +0.5 & +0.7 & +1.1 \\
\bottomrule
\end{tabular}
\end{table}

The consistency score (\Cref{tab:consistency}) is a neighborhood-level measure: it improves with fusion at every $K$ and at both scales, but it does not directly characterize global geometry. To complement it, we measured three global quantities on ImageNet embeddings for ITO with and without fusion across three pretraining scales: the image--text centroid gap (cosine similarity and L2 distance between the two modality centroids, following~\cite{liang2022mind}), the cross-modal MMD, and the ratio of same-class to different-class cross-modal distances, with standard deviations estimated from 50 bootstrap resamples. The effect of fusion on these quantities is small and not directionally consistent across scales. Fusion narrows the centroid gap at CC3M (centroid cosine $0.423{\to}0.466$) but widens it at DataComp-1B ($0.336{\to}0.314$). MMD and the class-distance ratio change only at the third to fourth decimal; with a bootstrap standard deviation of about $3{\times}10^{-4}$, these changes are genuinely small rather than measurement noise. No single global quantity tracks the downstream gains, which is why we rest the justification of fusion on the optimization evidence in~\Cref{align_integration}.

\section{UMAP Visualization}
\begin{figure}[htbp]
  \centering
  \begin{subfigure}[b]{0.32\columnwidth}
    \includegraphics[width=\linewidth]{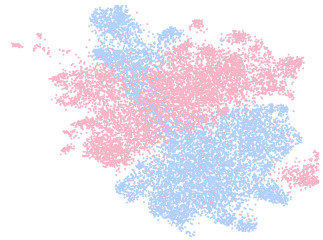}
    \caption{CLIP}
    \label{fig:clip}
  \end{subfigure}
  \hfill
  \begin{subfigure}[b]{0.32\columnwidth}
    \includegraphics[width=\linewidth]{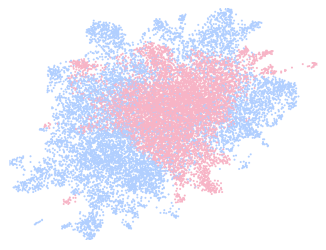}
    \caption{FLAIR}
    \label{fig:flair}
  \end{subfigure}
  \hfill
  \begin{subfigure}[b]{0.32\columnwidth}
    \includegraphics[width=\linewidth]{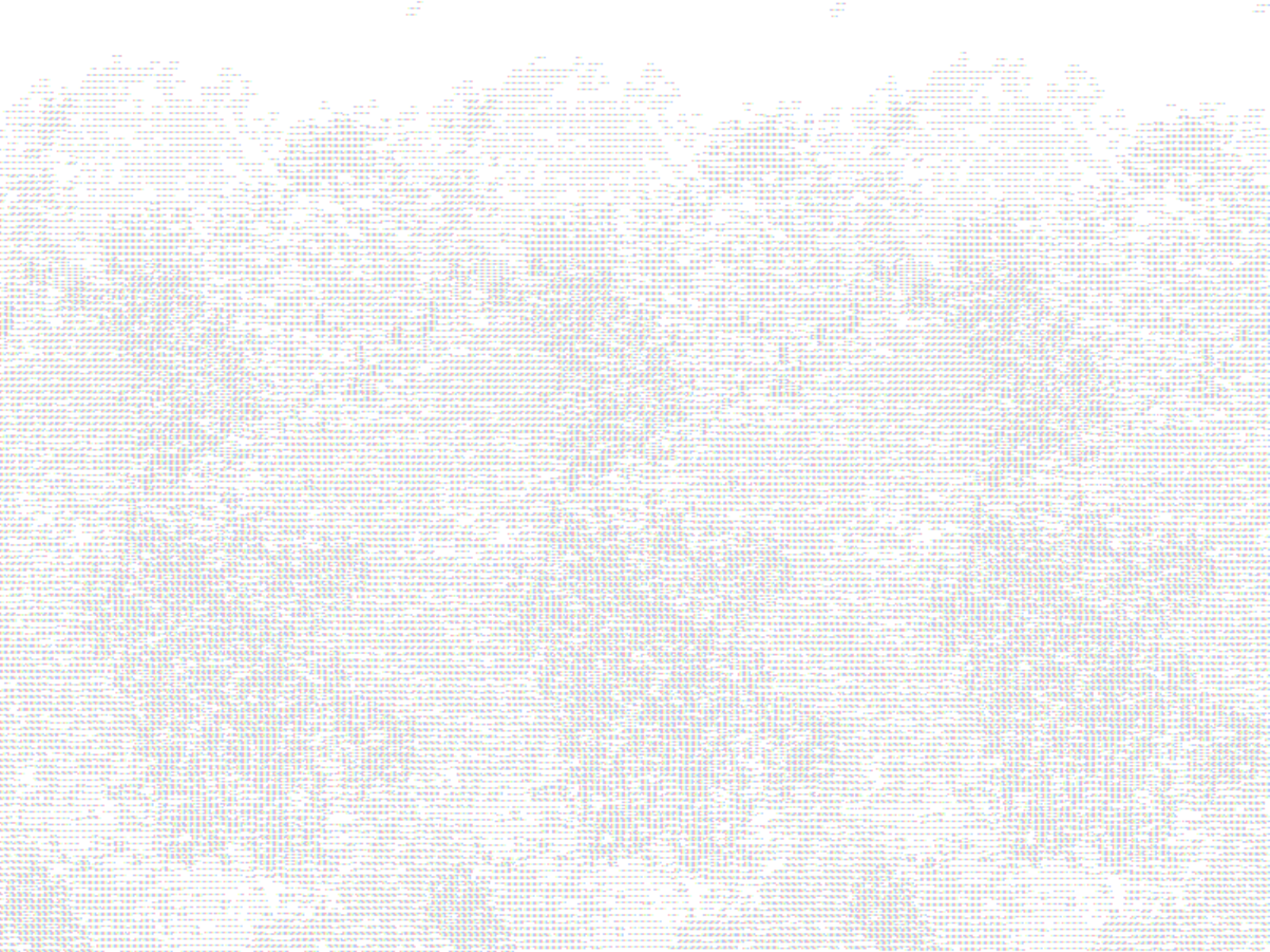}
    \caption{\textbf{ITO}}
    \label{fig:ITO}
  \end{subfigure}

  \caption{\textbf{UMAP visualization.} All models are trained on the CC3M dataset. For visualization, 8,192 image-text pairs are randomly sampled from CC12M. \textcolor{myblue}{Blue} points represent images, and \textcolor{myred}{red} points represent texts. (a): CLIP exhibits a clear separation between modalities, with a distinct boundary between image and text embeddings. (b): FLAIR shows more compact text embeddings surrounded by image embeddings, likely due to its text-conditioned fusion mechanism. (c): \name~shows a star-shaped distribution in which image and text embeddings are more closely interleaved and the boundary between modalities is less pronounced in this projection.}
  \label{fig:visual}
\end{figure}
\subsection{CLIP vs. FLAIR vs. ITO: Modality Separation vs. Integration}
\label{umap}
UMAP projections are qualitative: a two-dimensional embedding can exaggerate or hide structure, and the global geometry probes in Appendix~\ref{app:geometry} show that the effect of fusion on global quantities is small and scale-dependent. The observations below should be read as illustrations rather than as evidence of a geometric mechanism.

We first compare the representation structures learned by CLIP, FLAIR, and ITO using UMAP visualization under identical settings.
All models are trained on the CC3M dataset.
For visualization, we randomly sample 8,192 image--text pairs from CC12M and project their embeddings into two dimensions using UMAP.
Blue points represent image embeddings, and red points represent text embeddings.

As shown in~\Cref{fig:clip}, CLIP exhibits a clear separation between image and text embeddings, forming two modality-specific clusters with a distinct boundary.
This observation is consistent with prior findings that instance-level contrastive alignment alone does not eliminate modality-induced organization in the embedding space.

\Cref{fig:flair} presents the visualization for FLAIR.
Compared with CLIP, FLAIR produces more compact text embeddings that are partially surrounded by image embeddings.
This behavior is likely attributed to its text-conditioned pooling mechanism, which introduces localized fusion effects during training.
However, the overall structure remains asymmetric, with text embeddings forming a concentrated region and images distributed more broadly, indicating that modality separation is alleviated but not fully resolved.

In contrast,~\Cref{fig:ITO} shows the visualization for ITO.
Image and text embeddings are more closely interwoven, forming a star-shaped distribution in which both modalities are mixed within shared neighborhoods.
Notably, this visualization is obtained using only the standalone dual-encoder at inference time, without any fusion module.
This suggests that the reduced modality separation is not an artifact of architectural unification, but rather reflects a change in the representation structure learned during training.

Taken together, in these projections ITO appears less structured by modality than CLIP and FLAIR.

\subsection{Effect of Training-Time Fusion ($\lambda=0$ vs. $\lambda>0$)}
\begin{figure}[htbp]
  \centering
  \begin{subfigure}[b]{0.32\columnwidth}
    \includegraphics[width=\linewidth]{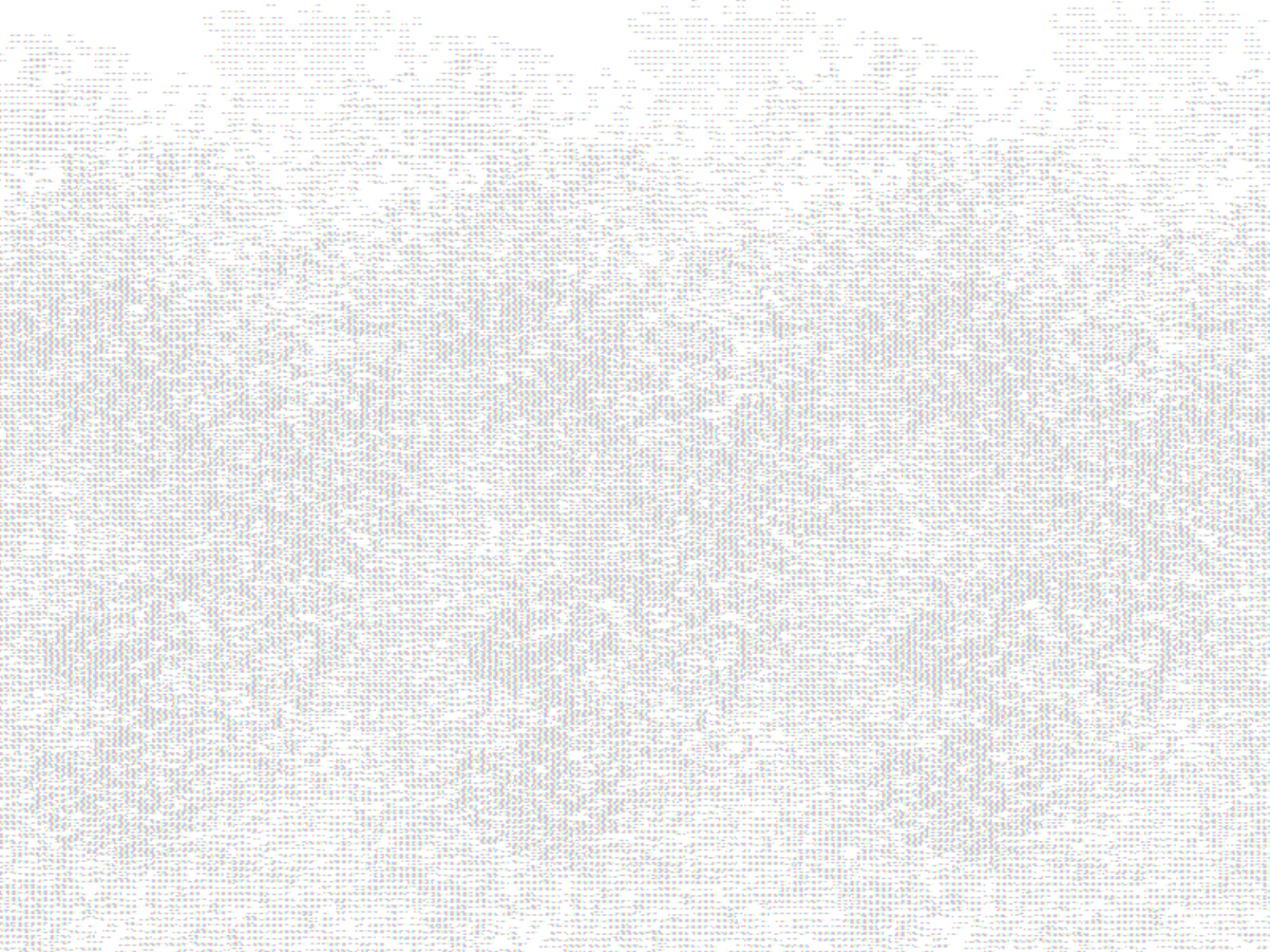}
    \caption{ITO ($\lambda=0$)}
    \label{fig:ITO_0}
  \end{subfigure}
  \hfill
  \begin{subfigure}[b]{0.32\columnwidth}
    \includegraphics[width=\linewidth]{figure/ito_cc3m_a2_lr3e3_visualization.pdf}
    \caption{ITO ($\lambda=2$)}
    \label{fig:ITO_2}
  \end{subfigure}
  \hfill
  \begin{subfigure}[b]{0.32\columnwidth}
    \includegraphics[width=\linewidth]{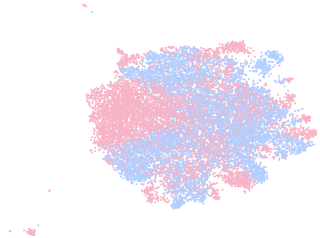}
    \caption{{ITO\_sub2}}
    \label{fig:ITO_sub2}
  \end{subfigure}

  \caption{\textbf{UMAP visualization.} All models are trained on the CC3M dataset. For visualization, 8,192 image-text pairs are randomly sampled from CC12M. \textcolor{myblue}{Blue} points represent images, and \textcolor{myred}{red} points represent texts.}
  \label{fig:visual2}
\end{figure}
To disentangle the effect of Multimodal Multiple Alignment from that of training-time multimodal fusion, we visualize the representation structures learned with different fusion weights.
All models are trained on CC3M, and 8,192 image--text pairs are randomly sampled from CC12M for UMAP visualization.
Inference is performed using the standalone dual-encoder in all cases.

\Cref{fig:ITO_0} shows the visualization for ITO with $\lambda=0$, where only Multimodal Multiple Alignment is applied.
Compared with CLIP, the separation between image and text embeddings is reduced, indicating improved instance-level alignment.
However, the embedding space remains partially organized by modality, and a visible boundary between image and text representations persists.

\Cref{fig:ITO_2} presents the full ITO model with $\lambda=2$.
With training-time multimodal fusion enabled, image and text embeddings appear more interwoven and the boundary between modalities is weaker in this projection on CC3M, where the centroid-gap probe also narrows (Appendix~\ref{app:geometry}).

\Cref{fig:ITO_sub2} further shows ITO\_sub2, which incorporates both training-time fusion and sub-description sampling.
The resulting embedding space exhibits the strongest degree of cross-modal mixing, with image and text representations densely interleaved within shared neighborhoods.
This is consistent with textual diversity being most useful in the moderate data regime, as discussed in~\Cref{sec:zs}.

\subsection{CC3M-recap: FLAIR vs. ITO\_sub2 vs. ITO\_sub3}
\begin{figure}[htbp]
  \centering
  \begin{subfigure}[b]{0.32\columnwidth}
    \includegraphics[width=\linewidth]{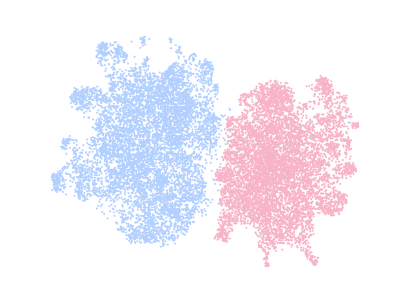}
    \caption{FLAIR}
  \end{subfigure}
  \hfill
  \begin{subfigure}[b]{0.32\columnwidth}
    \includegraphics[width=\linewidth]{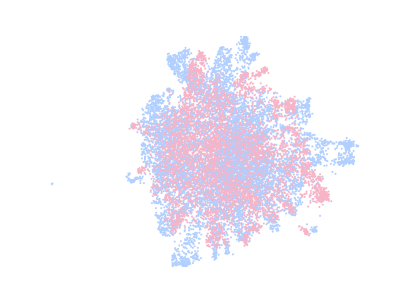}
    \caption{ITO\_sub2}
  \end{subfigure}
  \hfill
  \begin{subfigure}[b]{0.32\columnwidth}
    \includegraphics[width=\linewidth]{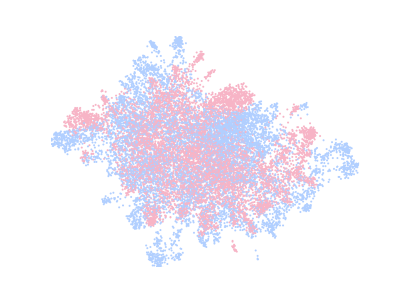}
    \caption{{ITO\_sub3}}
    \label{fig:ITO_sub3_recap}
  \end{subfigure}

  \caption{\textbf{UMAP visualization.} All models are trained on the CC3M-recap dataset. For visualization, 8,192 image-text pairs are randomly sampled from CC12M. \textcolor{myblue}{Blue} points represent images, and \textcolor{myred}{red} points represent texts.}
  \label{fig:ab-visual}
\end{figure}

We conduct additional visualization analyses to further examine the behavior of our model across different fusion strategies and caption granularities.
We first compare FLAIR~\cite{xiao2024flair}, \name\_sub2, and \name\_sub3 trained on the CC3M-recap dataset.
As shown in~\Cref{fig:ab-visual}, FLAIR remains partially organized by modality on CC3M-recap, exhibiting limited cross-modal mixing, in contrast to its slightly more compact structure observed on the original CC3M dataset.
This difference may be attributed to the fact that all subclauses in CC3M-recap originate from the same source caption, reducing the diversity of textual supervision.

In contrast, both \name\_sub2 and \name\_sub3 produce embedding spaces where image and text representations are more tightly interleaved.
In these projections, representations appear less separated by modality than those of FLAIR.

\subsection{DataComp-1B: CLIP vs. ITO}
\begin{figure}[htbp]
  \centering
  \begin{subfigure}[b]{0.45\columnwidth}
    \includegraphics[width=\linewidth]{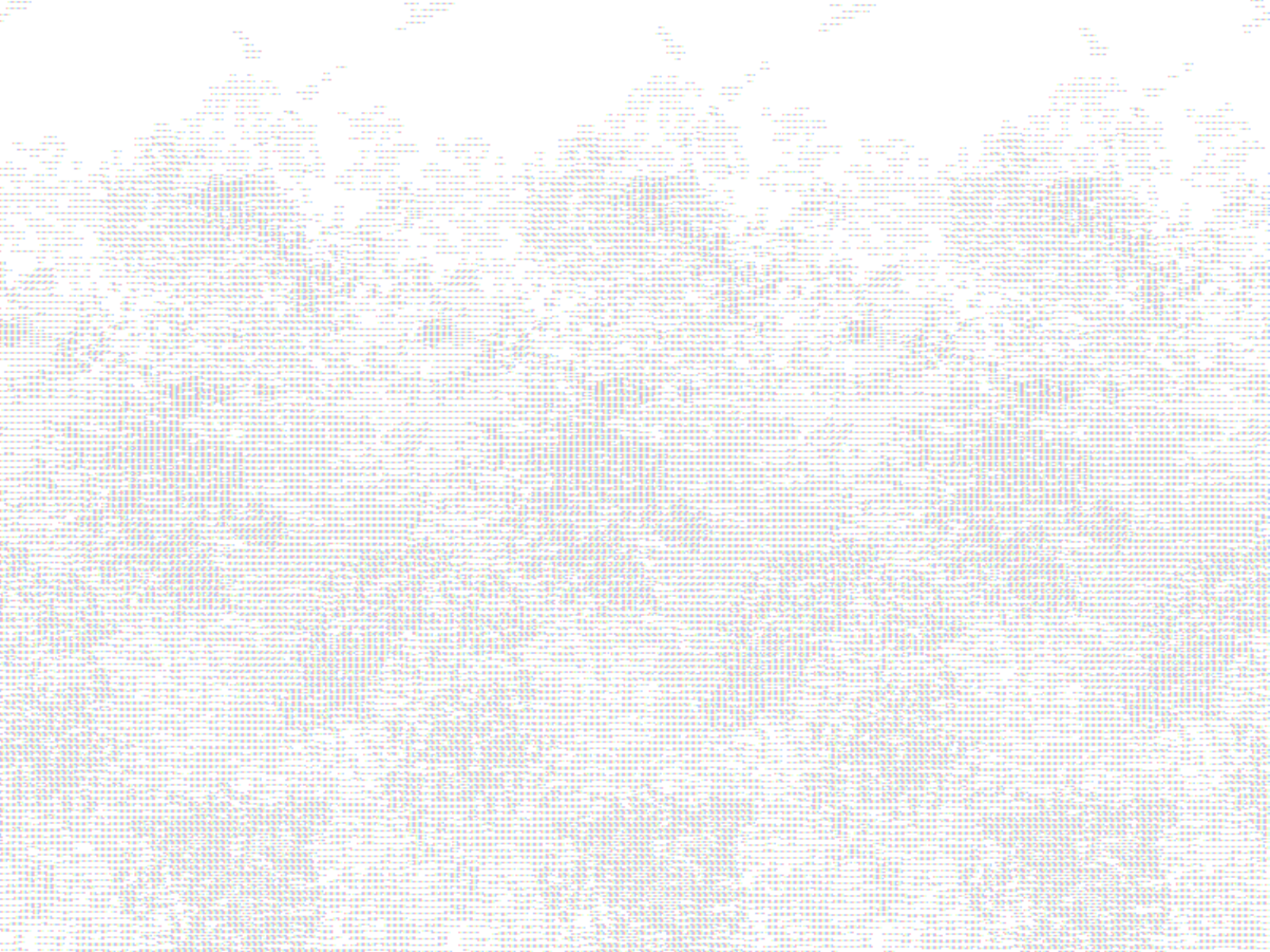}
    \caption{CLIP}
  \end{subfigure}
  \hfill
  \begin{subfigure}[b]{0.45\columnwidth}
    \includegraphics[width=\linewidth]{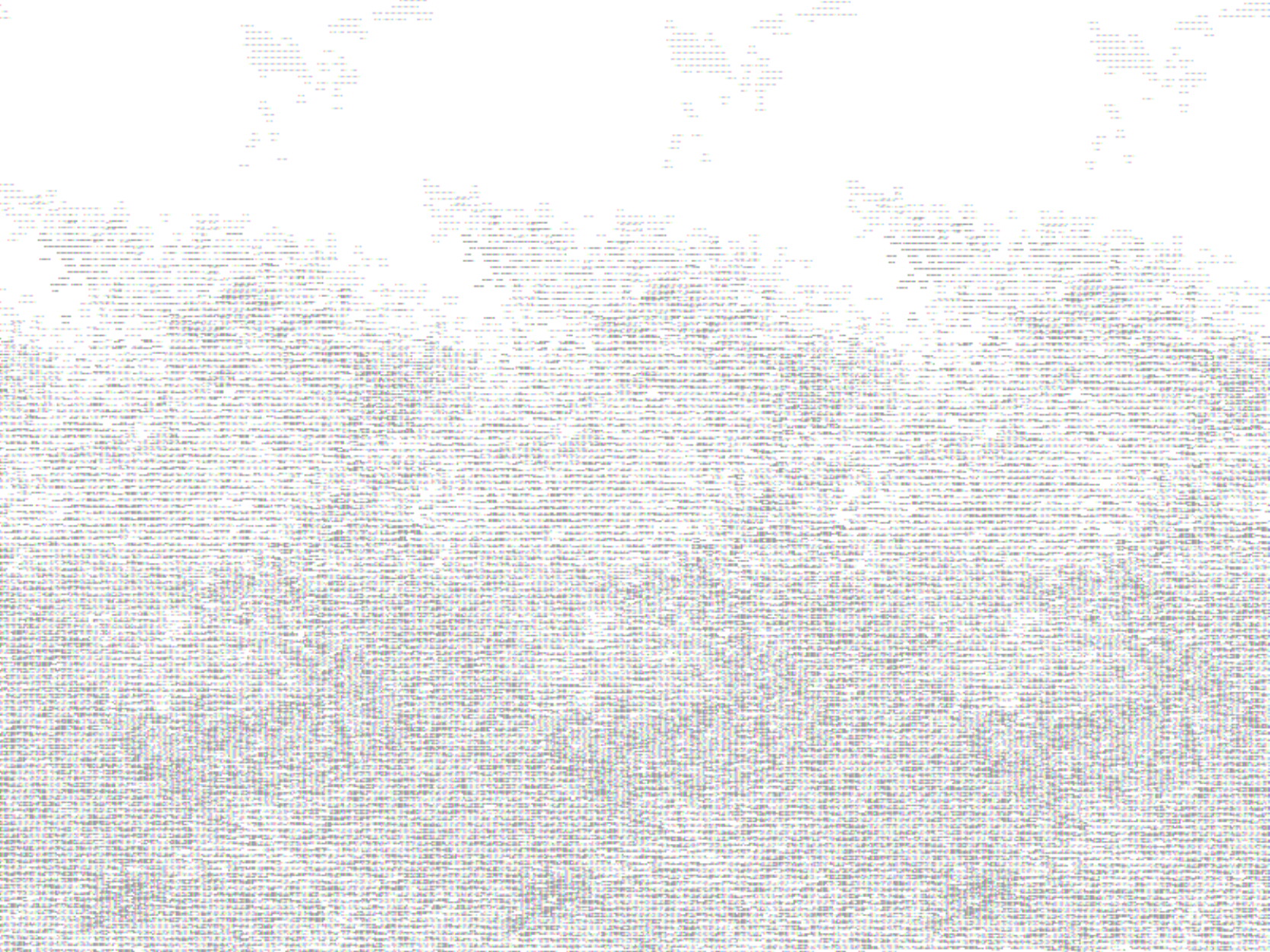}
    \caption{ITO}
  \end{subfigure}
  \hfill

  \caption{\textbf{UMAP visualization for DataComp-1B.} All models are trained on the DataComp-1B dataset for 10 epochs. For visualization, 8,192 image-text pairs are randomly sampled from CC12M. \textcolor{myblue}{Blue} points represent images, \textcolor{myred}{red} points represent texts, and \textcolor{mygreen}{green} points denote unified multimodal tokens.
.}
  \label{fig:ab-datacomp-1b}
\end{figure}
To investigate scalability, we further visualize models pretrained on the billion-scale DataComp-1B dataset.
Specifically, we compare CLIP~\cite{clip} and \name~using ViT-B/16, with both models pretrained for 10 epochs.
For visualization, 8,192 image--text pairs are randomly sampled from CC12M to ensure stable and consistent comparison across different pretraining scales.
Blue points represent image embeddings, red points denote text embeddings, and green points correspond to unified multimodal tokens.

As shown in~\Cref{fig:ab-datacomp-1b}, CLIP exhibits a clear modality-induced separation, with image and text embeddings forming two distinct clusters.
In contrast, \name~produces more intertwined distributions in which representations appear less structured by modality.
Note that this comparison is between CLIP and the full ITO; isolating the fusion module at this scale ($\lambda{=}0$ vs.\ $\lambda{=}2$), the centroid-gap probe does not narrow (Appendix~\ref{app:geometry}), so we do not attribute this visual difference to fusion alone.

\section{Training Dynamics and Overfitting Analysis}
\label{appendix:dynamics}
\begin{figure}[t!]
    \centering
    \includegraphics[width=0.6\linewidth]{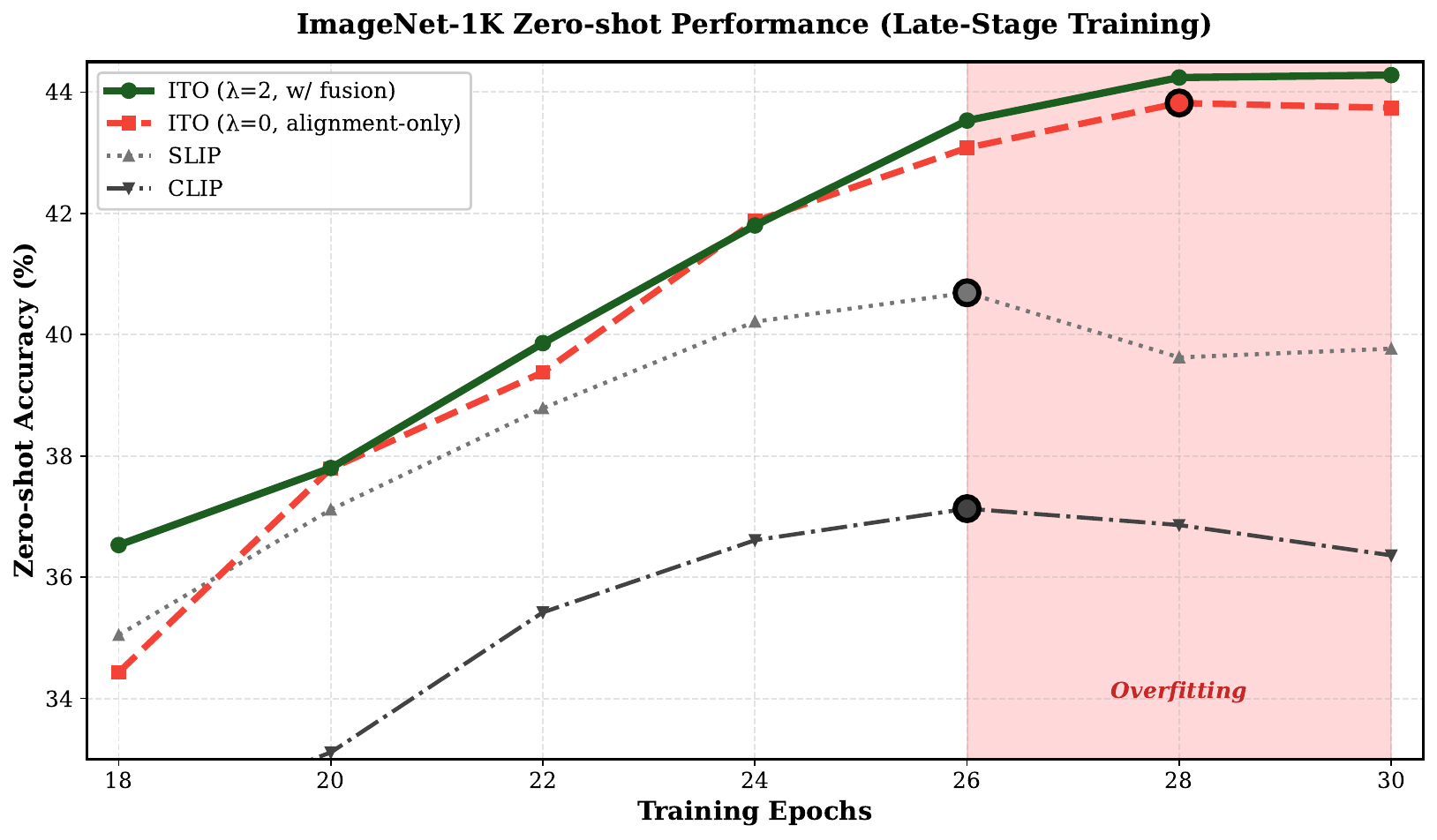}
    \caption{\textbf{Training dynamics and overfitting behavior on YFCC15M.}
Zero-shot ImageNet-1K accuracy as a function of training epochs for CLIP, SLIP, and ITO variants using ViT-B/16.
Both CLIP and SLIP exhibit late-stage performance degradation, indicating overfitting on moderately sized datasets.
Introducing Multimodal Multiple Alignment alone ($\lambda{=}0$) improves overall accuracy but does not fully prevent overfitting.
In contrast, enabling training-time multimodal fusion ($\lambda{=}2$) stabilizes training and maintains consistent zero-shot performance in later epochs, demonstrating the regularizing effect of fusion-based cross-modal interaction.}
    \label{fig:training_curve}
\end{figure}
To better understand the role of training-time multimodal fusion in image--text contrastive pretraining, we analyze the training dynamics of different methods by tracking zero-shot ImageNet-1K accuracy throughout training. All models are trained on YFCC15M using ViT-B/16, and evaluated under the standard CLIP zero-shot protocol. 

~\Cref{fig:training_curve} reports the evolution of zero-shot accuracy for CLIP, SLIP, and our proposed ITO variants. Both CLIP and SLIP exhibit a clear late-stage performance degradation: after reaching peak accuracy, their zero-shot performance declines as training proceeds, indicating overfitting despite continued optimization. This behavior is consistent with prior observations in image--text contrastive learning on moderately sized datasets.

Introducing Multimodal Multiple Alignment alone ($\lambda{=}0$) substantially improves overall accuracy compared with CLIP and SLIP. However, this alignment-only variant still suffers from noticeable late-stage overfitting, as reflected by a similar peak-and-decline pattern in the training curve. This suggests that enriching instance-level contrastive supervision, while beneficial, is insufficient to fully stabilize training. 

In contrast, enabling training-time multimodal fusion ($\lambda{=}2$) fundamentally changes the training dynamics. ITO exhibits consistently stable zero-shot performance in the later stages of training, without observable degradation. This indicates that the fusion objective provides an additional regularizing effect, guiding the encoders toward representations that generalize better and remain robust as training progresses. 

Taken together, since the $\lambda{=}0$ and $\lambda{=}2$ variants differ only in the fusion module, these results isolate overfitting suppression as a direct effect of fusion rather than of multi-view alignment. Importantly, these benefits are achieved without introducing any additional inference-time cost, as the fusion module is discarded during deployment.
\begin{figure*}[t!]
  \centering
  \begin{minipage}[t]{0.49\linewidth}
    \centering
    \subcaptionbox{CLIP\label{fig:clip1}}{%
      \includegraphics[width=\linewidth]{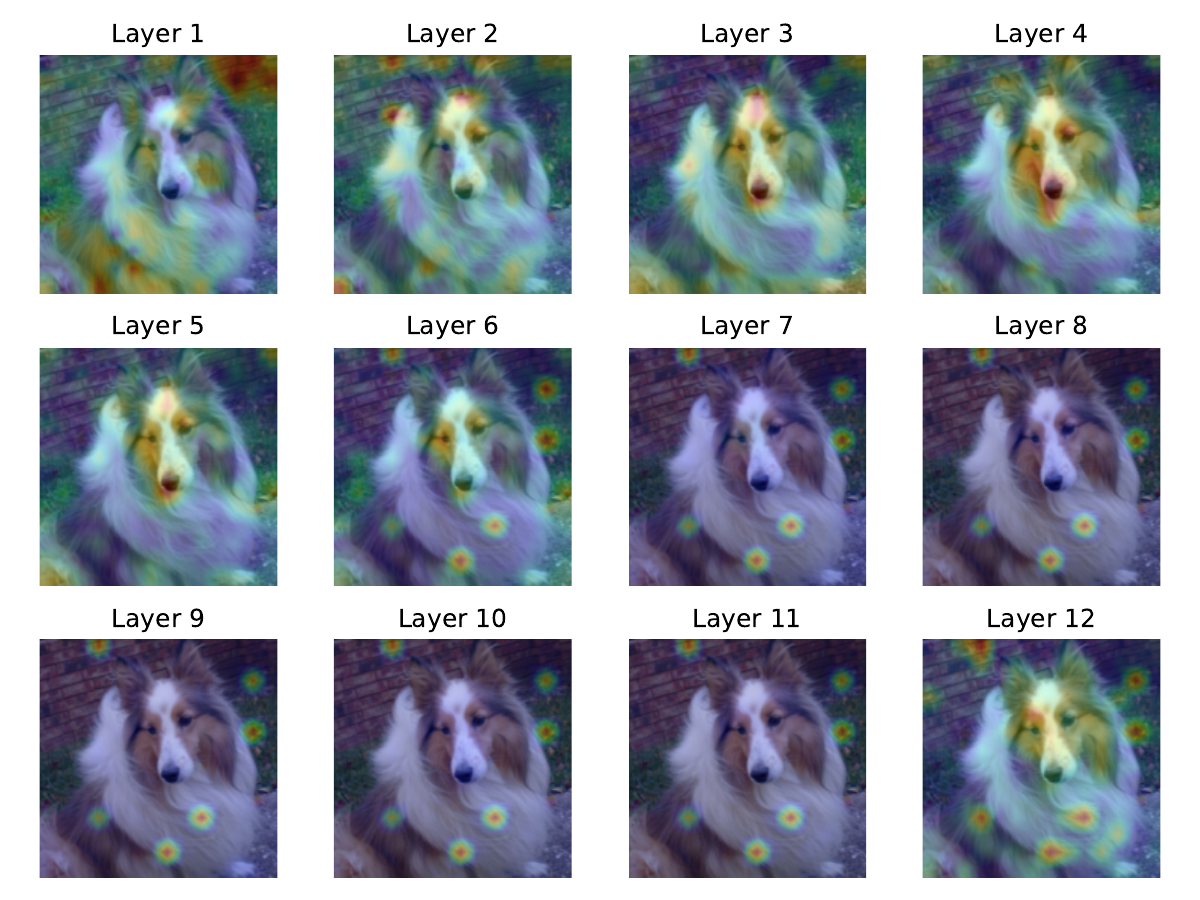}
    }\\[4pt]
    \subcaptionbox{CLIP\label{fig:clip2}}{%
      \includegraphics[width=\linewidth]{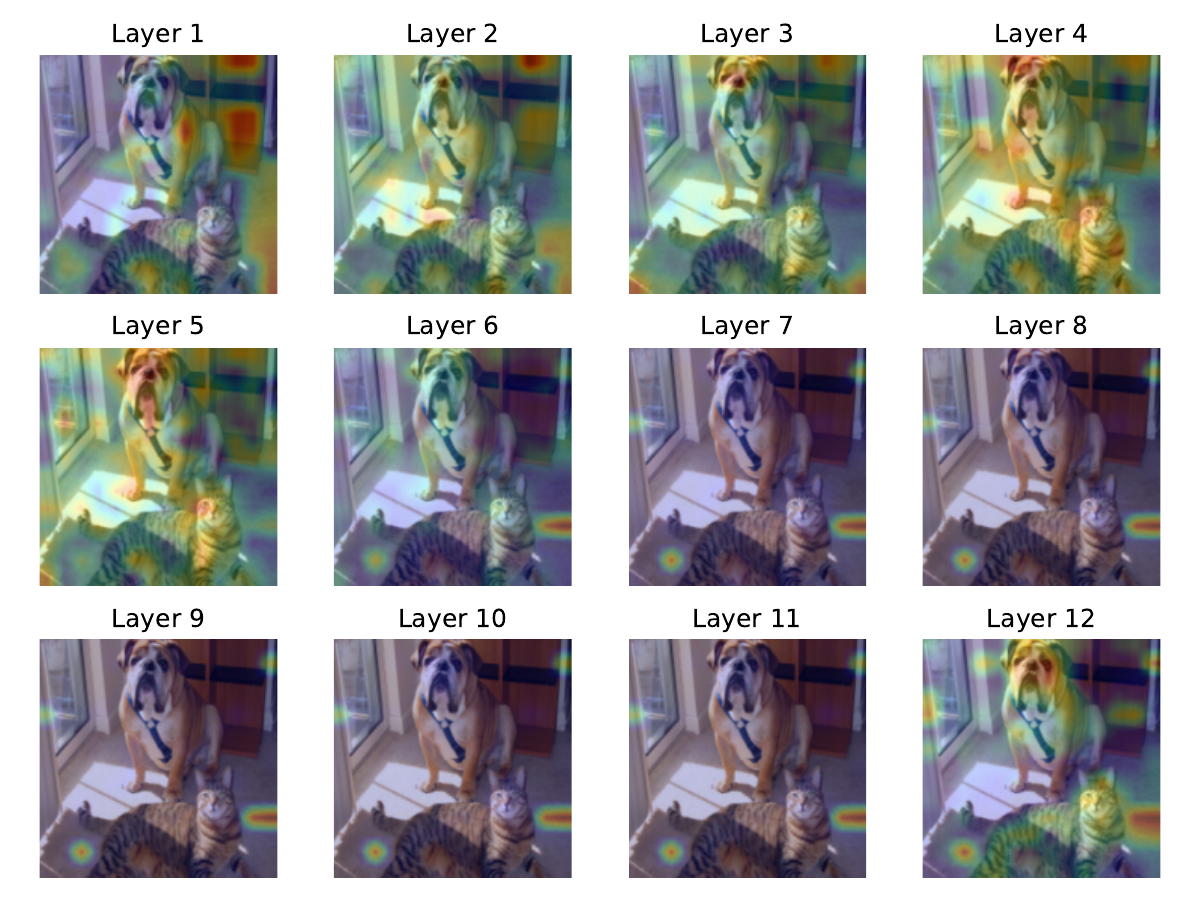}
    }
  \end{minipage}
  \hfill
  \begin{minipage}[t]{0.49\linewidth}
    \centering
    \subcaptionbox{ITO\label{fig:ito1}}{%
      \includegraphics[width=\linewidth]{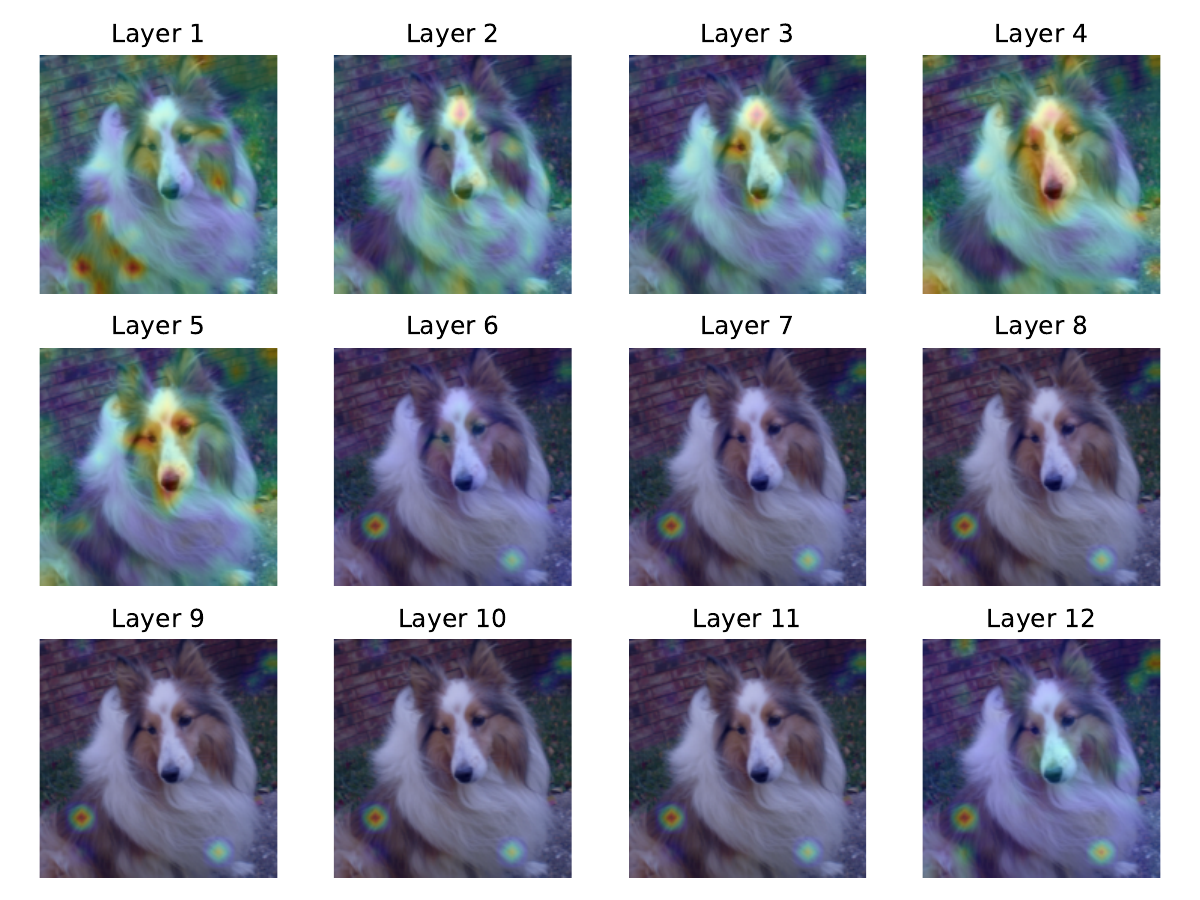}
    }\\[4pt]
    \subcaptionbox{ITO\label{fig:ito2}}{%
      \includegraphics[width=\linewidth]{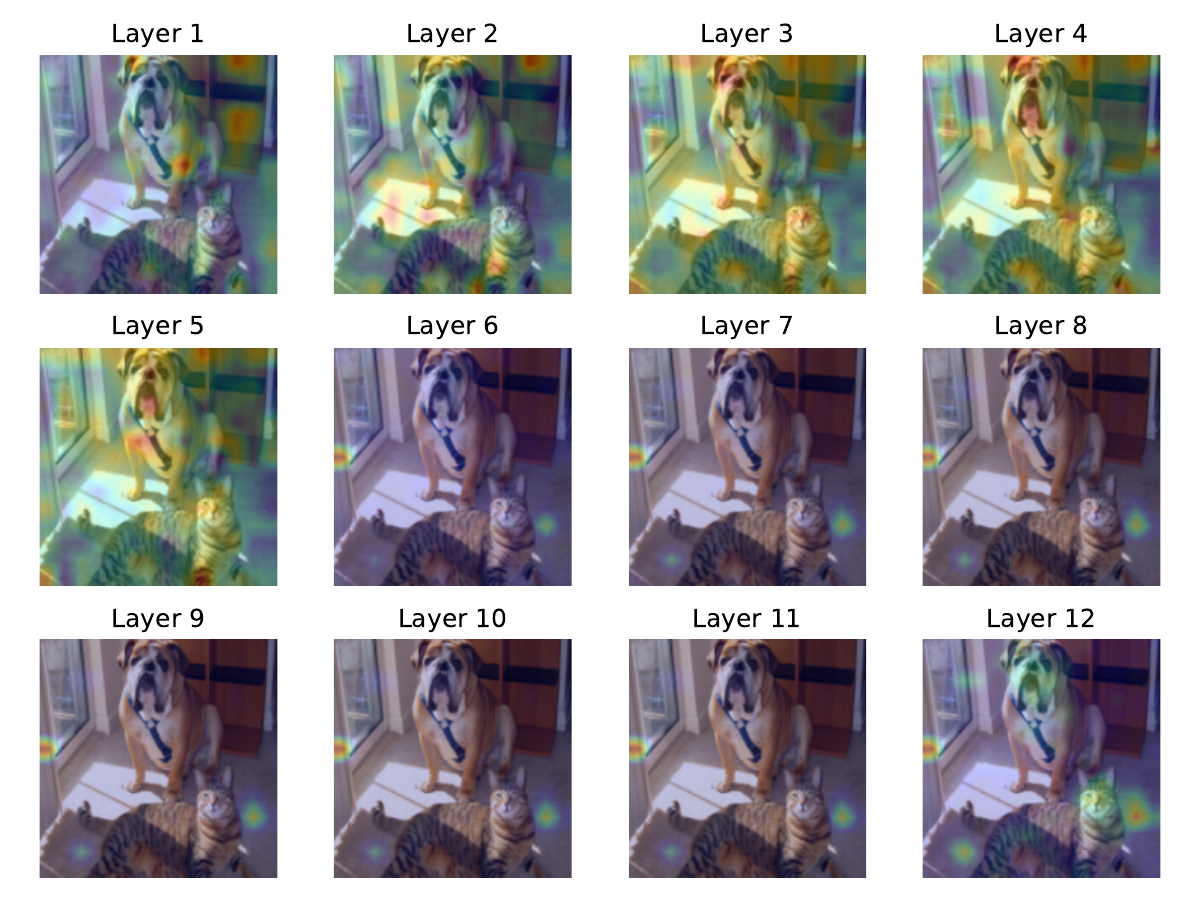}
    }
  \end{minipage}
  
  \caption{ \textbf{Layer-wise attention visualization of CLIP and \name~on DataComp-1B.} Each image corresponds to a different transformer layer, with the layer index indicated above each visualization. \name~shows progressively more focused and semantically consistent attention distributions than CLIP. Note that all visualizations are derived from held-out samples rather than those seen during pretraining. }
  \label{fig:clip-ito-4up}
\end{figure*}
\section{Layer-wise attention visualization}
We visualize the attention maps across all 12 layers of the vision encoders from CLIP~\cite{clip} and our \name~(trained for 10 epochs on DataComp-1B) to analyze their visual reasoning behaviors. 
As shown in~\Cref{fig:clip-ito-4up}, \name~exhibits more concentrated and semantically coherent attention distributions across layers. 
From shallow to deep layers, the attention in \name~progressively evolves from dispersed low-level responses to high-level representations that focus on the semantic subject of the scene, while effectively suppressing background noise. 
This indicates that \name~forms a clearer hierarchical structure for semantic information aggregation and achieves stronger alignment between visual features and semantic content. 
In contrast, CLIP shows less stable attention evolution, with certain layers focusing on irrelevant background regions and exhibiting weaker inter-layer consistency. 
Overall, the representations learned by \name~are more structured and interpretable, demonstrating enhanced semantic consistency and better generalization capability.

\section{Inference Efficiency.}
ITO introduces a multimodal fusion module \emph{only during training}.
At inference time, the fusion module is entirely removed, and ITO reduces
to a standard dual-encoder architecture identical to CLIP.
As a result, ITO has the same number of parameters, computational cost,
and inference latency as CLIP, and can be used as a drop-in replacement
for existing image--text contrastive encoders.

\section{Training Overhead Analysis}
\label{time-analysis}
\Cref{tab:cost} in the main paper summarizes the training cost of ITO; \Cref{tab:cost_full} gives the full breakdown, including the alignment-only variant. All measurements use 64 NVIDIA A800 GPUs with matched batch size and optimizer settings.

\begin{table}[h]
\centering
\caption{Training time (hours) and peak GPU memory (GB per GPU) of OpenCLIP and ITO. Ratios are relative to OpenCLIP at the same scale.}
\label{tab:cost_full}
\setlength{\tabcolsep}{6pt}
\renewcommand{\arraystretch}{1.05}
\small
\begin{tabular}{llcccc}
\toprule
Dataset & Method & Time (h) & Time ratio & Memory (GB) & Memory ratio \\
\midrule
\multirow{3}{*}{YFCC15M}
  & OpenCLIP            & 3.6        & 1.0$\times$ & 7.9  & 1.0$\times$ \\
  & ITO ($\lambda{=}0$) & 8.2        & 2.3$\times$ & 16.8 & 2.1$\times$ \\
  & ITO ($\lambda{=}2$) & 8.2        & 2.3$\times$ & 16.8 & 2.1$\times$ \\
\midrule
\multirow{3}{*}{DataComp-1B}
  & OpenCLIP            & 8.1        & 1.0$\times$ & 30.0 & 1.0$\times$ \\
  & ITO ($\lambda{=}0$) & ${\sim}18$ & 2.2$\times$ & 68.6 & 2.3$\times$ \\
  & ITO ($\lambda{=}2$) & ${\sim}18$ & 2.2$\times$ & 68.6 & 2.3$\times$ \\
\bottomrule
\end{tabular}
\end{table}

Two observations are worth noting. First, the overhead introduced by the fusion regularizer itself is negligible: ITO with $\lambda{=}0$ (alignment only) and $\lambda{=}2$ (alignment plus fusion) have indistinguishable cost at both scales. The training overhead almost entirely comes from the multi-view alignment pipeline. Second, this training-time cost does not propagate to inference: ITO has identical parameter count, FLOPs, and latency to a standard CLIP dual-encoder at deployment.


\newpage
\section*{NeurIPS Paper Checklist}

\begin{enumerate}

\item {\bf Claims}
    \item[] Question: Do the main claims made in the abstract and introduction accurately reflect the paper's contributions and scope?
    \item[] Answer: \answerYes{} 
    \item[] Justification: We summarize the contributions of the paper in the introduction.

\item {\bf Limitations}
    \item[] Question: Does the paper discuss the limitations of the work performed by the authors?
    \item[] Answer: \answerYes{} 
    \item[] Justification: The Limitations paragraph in~\Cref{sec:con} discusses the training-time overhead (\Cref{tab:cost}), the limited set of baselines at 100M--1B scale, the modest accuracy gain and indirect geometric evidence for fusion, the sensitivity to augmentation and text-view construction, and the absence of a data-quality ablation.

\item {\bf Theory assumptions and proofs}
    \item[] Question: For each theoretical result, does the paper provide the full set of assumptions and a complete (and correct) proof?
    \item[] Answer: \answerNA{} 
    \item[] Justification: There are no theoretical results in this paper.

    \item {\bf Experimental result reproducibility}
    \item[] Question: Does the paper fully disclose all the information needed to reproduce the main experimental results of the paper to the extent that it affects the main claims and/or conclusions of the paper (regardless of whether the code and data are provided or not)?
    \item[] Answer: \answerYes{} 
    \item[] Justification: In~\Cref{settings}, we have provided the detailed experimental settings for results reproducing.

\item {\bf Open access to data and code}
    \item[] Question: Does the paper provide open access to the data and code, with sufficient instructions to faithfully reproduce the main experimental results, as described in supplemental material?
    \item[] Answer:  \answerYes{} 
    \item[] Justification: Code is publicly available at \url{https://github.com/showstarpro/ITO}. All pretraining datasets are public, and the experimental setup is detailed in~\Cref{settings} and Appendix~\ref{app:aug}.

\item {\bf Experimental setting/details}
    \item[] Question: Does the paper specify all the training and test details (e.g., data splits, hyperparameters, how they were chosen, type of optimizer) necessary to understand the results?
    \item[] Answer: \answerYes{} 
    \item[] Justification: We have provided the detailed experimental settings in~\Cref{settings}.

\item {\bf Experiment statistical significance}
    \item[] Question: Does the paper report error bars suitably and correctly defined or other appropriate information about the statistical significance of the experiments?
    \item[] Answer: \answerNo{} 
    \item[] Justification: Each configuration is trained once, since repeating large-scale image--text pretraining over multiple random seeds is computationally prohibitive (up to billion-scale data). We follow the standard evaluation protocols of prior work and report results over many benchmarks (26 zero-shot classification datasets, retrieval on three datasets, and 13 multimodal benchmarks) and multiple pretraining scales to reduce the dependence on any single measurement.

\item {\bf Experiments compute resources}
    \item[] Question: For each experiment, does the paper provide sufficient information on the computer resources (type of compute workers, memory, time of execution) needed to reproduce the experiments?
    \item[] Answer: \answerYes{} 
    \item[] Justification: We describe the hardware setup, software environment, and codebase in~\Cref{settings}, and report training time and peak GPU memory in~\Cref{tab:cost} and Appendix~\ref{time-analysis}.
    
\item {\bf Code of ethics}
    \item[] Question: Does the research conducted in the paper conform, in every respect, with the NeurIPS Code of Ethics \url{https://neurips.cc/public/EthicsGuidelines}?
    \item[] Answer: \answerYes{} 
    \item[] Justification: Our paper is conducted with the NeurIPS Code of Ethics.

\item {\bf Broader impacts}
    \item[] Question: Does the paper discuss both potential positive societal impacts and negative societal impacts of the work performed?
    \item[] Answer: \answerNA{} 
    \item[] Justification: This is foundational research on training objectives for image--text contrastive pretraining, trained on public datasets and not tied to a particular application or deployment, so we do not discuss societal impacts specific to this work.
    
\item {\bf Safeguards}
    \item[] Question: Does the paper describe safeguards that have been put in place for responsible release of data or models that have a high risk for misuse (e.g., pre-trained language models, image generators, or scraped datasets)?
    \item[] Answer: \answerNA{} 
    \item[] Justification: This paper does not release generative models. And this paper poses no such risks.

\item {\bf Licenses for existing assets}
    \item[] Question: Are the creators or original owners of assets (e.g., code, data, models), used in the paper, properly credited and are the license and terms of use explicitly mentioned and properly respected?
    \item[] Answer: \answerYes{} 
    \item[] Justification: We conduct experiments in this paper using publicly available datasets and codebases, and cite relevant references.

\item {\bf New assets}
    \item[] Question: Are new assets introduced in the paper well documented and is the documentation provided alongside the assets?
   \item[] Answer: \answerNA{} 
    \item[] Justification: This paper does not release new assets.

\item {\bf Crowdsourcing and research with human subjects}
    \item[] Question: For crowdsourcing experiments and research with human subjects, does the paper include the full text of instructions given to participants and screenshots, if applicable, as well as details about compensation (if any)? 
    \item[] Answer: \answerNA{} 
    \item[] Justification: This paper does not involve crowdsourcing nor research with human subjects.

\item {\bf Institutional review board (IRB) approvals or equivalent for research with human subjects}
    \item[] Question: Does the paper describe potential risks incurred by study participants, whether such risks were disclosed to the subjects, and whether Institutional Review Board (IRB) approvals (or an equivalent approval/review based on the requirements of your country or institution) were obtained?
    \item[] Answer: \answerNA{} 
    \item[] Justification: This paper does not involve crowdsourcing nor research with human subjects.

\item {\bf Declaration of LLM usage}
    \item[] Question: Does the paper describe the usage of LLMs if it is an important, original, or non-standard component of the core methods in this research? Note that if the LLM is used only for writing, editing, or formatting purposes and does \emph{not} impact the core methodology, scientific rigor, or originality of the research, declaration is not required.
    \item[] Answer: \answerNA{} 
    \item[] Justification: The core method development in this research does not involve LLMs as any important, original, or non-standard components.

\end{enumerate}

\end{document}